\documentclass[sigconf]{acmart}
\AtBeginDocument{%
  }

\usepackage{graphicx}
\usepackage{url}
\usepackage{booktabs}
\usepackage{diagbox}
\usepackage{bm}
\usepackage{amsmath}

\usepackage{pifont}
\usepackage[]{algpseudocode}
\usepackage{algorithm}
\algrenewcommand\textproc{\texttt}
\usepackage{caption}
\usepackage{bigstrut}
\usepackage{subfig}
\usepackage{enumitem}
\usepackage{float}
\usepackage{balance}
\usepackage{multirow}
\usepackage{tabularx}
\usepackage{pdfpages}
\usepackage{microtype}
\usepackage{mathtools}
\usepackage{threeparttable}
\usepackage{makecell}

\newcommand{\textoverline}[1]{$\overline{\mbox{#1}}$}

\setcopyright{none}
\copyrightyear{2027}
\acmYear{2027}
\acmDOI{XXXXXXX.XXXXXXX}

\begin{document}

\title{OpenGLT: A Comprehensive Benchmark of Graph Neural Networks for Graph-Level Tasks  [Experiments \& Analysis]}
 
\title{OpenGLT: A Comprehensive Benchmark of Graph Neural Networks for Graph-Level Tasks}
\settopmatter{authorsperrow=4}
\author{Haoyang Li}
\affiliation{%
	\institution{COMP, PolyU}
	\city{Hong Kong SAR}
	\country{China}
}
\email{haoyang-comp.li@polyu.edu.hk}

\author{Yuming Xu}
\affiliation{%
	\institution{COMP, PolyU}
	\city{Hong Kong SAR}
	\country{China}
}
\email{martin.xu@connect.polyu.hk}

\author{Alexander Zhou}
\affiliation{%
	\institution{COMP, PolyU}
	\city{Hong Kong SAR}
	\country{China}
}
\email{alexander.zhou@polyu.edu.hk}

\author{Yongqi Zhang}
\affiliation{%
	\institution{DSA, HKSUT (GZ)}
	\city{Guangzhou}
	\country{China}
}
\email{yongqizhang@hkust-gz.edu.cn}

\author{Jason Chen Zhang}
\affiliation{%
	\institution{COMP \& SHTM , PolyU}
\city{Hong Kong SAR}
\country{China}
}
\email{jason-c.zhang@polyu.edu.hk}

\author{Lei Chen}
\affiliation{%
	\institution{DSA, HKSUT (GZ)}
	\city{Guangzhou}
	\country{China}
}
\email{leichen@cse.ust.hk}

\author{Qing Li}
\affiliation{%
	\institution{COMP, PolyU}
	\city{Hong Kong SAR}
	\country{China}
}
\email{csqli@comp.polyu.edu.hk}

\renewcommand{\shortauthors}{Li, et al.}

\begin{abstract}

Graphs are fundamental data structures for modeling complex interactions in domains such as social networks, molecular structures, and biological systems. Graph-level tasks, which involve predicting properties or labels for entire graphs, are crucial for applications like molecular property prediction and subgraph counting. While Graph Neural Networks (GNNs) have shown significant promise for these tasks, their evaluations are often limited by narrow datasets, insufficient architecture coverage, restricted task scope and scenarios, and inconsistent experimental setups, making it difficult to draw reliable conclusions across domains. In this paper, we present a comprehensive experimental study of GNNs on graph-level tasks, systematically categorizing them into five types: node-based, hierarchical pooling-based, subgraph-based, graph learning-based, and self-supervised learning-based GNNs. We propose a unified evaluation framework OpenGLT, which standardizes evaluation across four domains (social networks, biology, chemistry, and motif counting), two task types (classification and regression), and three real-world scenarios (clean, noisy, imbalanced, and few-shot graphs). Extensive experiments on 20 models across 26 classification and regression datasets reveal that:  (i) no single architecture dominates both effectiveness and efficiency universally, i.e., subgraph-based GNNs excel in expressiveness, graph learning-based and SSL-based methods in robustness, and node-based and pooling-based models in efficiency; and (ii) specific graph topological features  such as density and centrality can partially guide the selection of suitable GNN architectures for different graph characteristics.

\end{abstract}


\maketitle

 \section{Introduction}\label{sec:intro}
Graphs represent objects as nodes and their relationships as edges, which are key data structures across various domains to model complex interactions~\cite{bonifati2024future,fan2022big,fang2022densest}, 
such as social networks, molecular structures, biological systems, etc.
Early graph representation learning methods~\cite{perozzi2014deepwalk,grover2016node2vec,wang2016structural,dong2017metapath2vec} laid the foundation of graph learning. 
Graph-level tasks aim at predicting properties of an entire graph, rather than individual nodes or edges. 
These tasks are crucial in domains including subgraph counting in database management~\cite{fichtenberger2022approximately,li2024fast}, molecular property prediction in chemistry~\cite{zhong2023knowledge,hu2020open}, and protein classification in bioinformatics~\cite{morris2020tudataset,li20242}.
Recently, graph neural networks (GNNs)~\cite{wang2023hongtu,xiang2025capsule,liu2025diskgnn,song2023adgnn,liao2022scara,cui2021metro,DBLP:journals/pvldb/LiJWSC24} have emerged as powerful tools for graph-structured and anomaly detection tasks~\cite{ma2023towards}. 
They learn node representations by iteratively aggregating neighbor information to obtain graph representations.

Depending on the approach,  we categorize GNNs for graph-level tasks into five categories: node-based, hierarchical pooling (HP)-based, subgraph-based, graph learning (GL)-based, and self-supervised learning (SSL)-based methods. 
Node-based GNNs~\cite{kipf2016semi,hamilton2017inductive,xu2018how,velivckovic2017graph} compute node representations through message passing and aggregate them via a permutation-invariant readout function, such as averaging, to form the final graph embedding. 
HP-based GNNs~\cite{mesquita2020rethinking,dhillon2007weighted,li2022cc,bianchi2020spectral} apply pooling operations to reduce the graph size and capture hierarchical structure, yielding multi-level graph representations. 
Subgraph-based GNNs~\cite{ding2024sgood,yan2024efficient,bevilacqua2024efficient,papp2022theoretical} divide the graph into subgraphs, learn a representation for each, and then aggregate these to represent the whole graph. 
GL-based GNNs~\cite{fatemi2023ugsl,zhiyao2024opengsl,li2024gslb,li2024fight} enhance graph quality by reconstructing structure and features. 
SSL-based GNNs~\cite{sun2024motif,wang2022graph,inae2023motif,li20242,thakoor2021bootstrapped} pretrain on unlabeled data, either by predicting graph properties or maximizing agreement between augmented views of the same graph.

Although various GNNs have been designed for graph-level tasks, their evaluations are often restricted to a narrow range of domain-specific datasets and insufficient baseline comparisons~\cite{zhiyao2024opengsl,errica2020fair,li2023gslb,wang2024comprehensive}. 
To ensure fair comparison, we identify and address five key shortcomings in current evaluation frameworks.
 
\begin{itemize}[leftmargin=*]
	\item \textbf{Issue 1. No clear taxonomy for GNNs on graph-level Tasks.}
	Graph-level tasks require different approaches than node-level tasks, yet a clear taxonomy for GNNs on graph-level tasks is lacking. This gap hinders holistic understanding and systematic comparison of models.
	
%
%
\item \textbf{Issue 2. Inconsistent Evaluation Pipelines.}
The lack of a standardized evaluation pipeline results in inconsistent comparisons.
Different works often use varying data splits, tuning protocols, and evaluation metrics, hindering fair assessment of model performance. A unified evaluation framework is needed for transparent and reliable comparisons.

\item \textbf{Issue 3. Restricted Coverage of GNN Architectures.}
Most evaluations focus on a limited set of architectures, such as node-based GNNs, while often ignoring more expressive models like subgraph-based GNNs. This narrow coverage limits performance comparisons and overlooks the strengths of diverse approaches.

\item \textbf{Issue 4. Insufficient Data Diversity.}
Current evaluations typically use datasets from a narrow range of domains, such as chemistry or biology. This limited diversity can lead to overfitting and restricts the generalizability of GNNs to other domains like social networks or different types of graphs.

	\item \textbf{Issue 5. Narrow Task and Scenario Scope.}
Current evaluation frameworks typically focus on a single type of graph-level task, such as molecular graph classification, and overlook diverse applications like cycle or path counting. They also assume access to ample clean labeled data, neglecting real-world challenges such as noise, class imbalance, or limited labeled graphs.


\end{itemize}
To address these five issues, we introduce OpenGLT, a comprehensive framework designed to provide a fair and thorough assessment of GNNs for graph-level tasks. To address the lack of a clear taxonomy, we systematically categorize existing GNNs for graph-level tasks into five distinct types and conduct an in-depth analysis of each type to understand their unique strengths and limitations. To address the inconsistent evaluation pipelines, we introduce a unified framework with standardized data splitting, tuning protocols, and evaluation metrics, ensuring fair and reproducible comparisons. To address the restricted architecture coverage, we include 20 representative models spanning all five categories, enabling comprehensive and systematic comparisons across diverse approaches. To address the insufficient data diversity, we incorporate graph datasets from diverse domains, including biology, chemistry, social networks, and motif graphs, ensuring broad and representative evaluations. To address the narrow task and scenario scope, we comprehensively evaluate GNNs on both graph classification and graph regression tasks, and further assess them under real-world scenarios, including noisy graphs, imbalanced datasets, and few-shot learning settings. Moreover, we conduct a correlation analysis between graph topological properties and model performance, providing practical guidance for architecture selection based on graph characteristics. 
We summarize the contributions as follows.

\begin{itemize}[leftmargin=10pt]
	\item We systematically revisit GNNs for graph-level tasks and categorize them into five types,  with in-depth analysis in Section~\ref{sec:method}.
 
	\item 
	We propose a unified open-source evaluation framework, OpenGLT, which covers diverse tasks, datasets, and scenarios.
	
	\item We conduct extensive experiments with 20  models across 26 datasets, complemented by a correlation analysis between graph properties and model performance, offering practical insights and architecture selection guidance.

\end{itemize}

\section{Preliminaries and Related Work }\label{sec:related_work}
We first introduce graph-level tasks and then review existing experimental studies.
Important notations  are in in Table~\ref{tab:notation}.

\noindent\textbf{Graph-level Tasks. }
Graph-level tasks, including classification and regression, are widely applied across domains, such as recommendation systems in social networks~\cite{cohen2016data,linghu2020global} with techniques like session contexts~\cite{wang2020global}, molecular property and activity prediction in chemistry~\cite{chen2024view,cohen2016data,mandal2022metalearning,niazi2023recent,gilmer2017neural}, and protein analysis, motif identification, or gene expression prediction in biology~\cite{zang2023hierarchical,otal2024analysis}.
Formally,  we denote a graph as 
$G_i(V_i, \mathbf{A}_i, \mathbf{X}_i)$,
where $V_i$, $\mathbf{A}_i \in \{0,1\}^{|V_i| \times |V_i|}$, $\mathbf{X}_i \in \mathbb{R}^{|V_i| \times d_x}$, denote nodes, adjacency matrix, and node features, respectively. 
In general, graph-level tasks aim to learn a mapping from $G_i$ to either a discrete label $y_i$ for classification (e.g., molecular property prediction~\cite{mandal2022metalearning,niazi2023recent} and query execution plan selection~\cite{zhao2022queryformer,zhou2020query}) or a continuous value for regression (e.g., motif counting~\cite{zhao2021learned,li2024fast} and cardinality estimation~\cite{GNCE,wang2022neural,teng2024cardinality}).

\noindent\textbf{Related Works of Experimental Studies}
As summarized in Table~\ref{tab:related_work}, existing benchmarks and surveys~\cite{zhiyao2024opengsl,li2024rethinking,hu2020open,errica2020fair,wang2024comprehensive, liao2025comprehensive} often face four key limitations~\cite{zhiyao2024opengsl,li2024rethinking,hu2020open,errica2020fair,wang2024comprehensive}: 
(i) a lack of systematic taxonomy for graph-level tasks; 
(ii) a predominant focus on node-based GNNs~\cite{dwivedi2023benchmarking, hu2020open,morris2020tudataset}, leaving diverse architectures partially absent with poor categorization; 
(iii) the neglect of realistic scenarios such as noise, imbalance, and few-shot settings, which hinders robustness assessment; 
and (iv) a lack of comprehensive efficiency metrics for usability evaluation.

 \begin{table}[t]
	\centering
	\small
	\vspace{0em}
	\caption{Summary on important notations.}
	\label{tab:notation}
	\setlength\tabcolsep{6pt}
	\begin{tabular}{l|p{6cm}}
		\toprule
		\bf Symbols & \bf Meanings \\ \midrule
		
		{$G_i(V_i, \mathbf{A}_i, \mathbf{X}_i)$ }    &    
		The graph $G_i$  \\ \hline
		
		$y_i$ &  Discrete label or continuous  value of $G_i$ \\ \hline
		
		$N^l_i{(v)}, N_i(v)$ & The $l$-hop and 1-hop neighbors of   $v$ in $G_i$    \\ \hline
		
		
		$f_\theta$ & GNN model \\ \hline
		
		$l, L$ & GNN layer index and the total layer number  \\ \hline
		
		$\mathbf{h}^{(l)}_i(v)$  & The $l$-th layer   representation of $v$ in $G_i$\\ \hline 
		
		
		$e_{uv}$ & Edge feature of edge $(u,v)$ in $G_i$ \\ \hline
		
		$\mathbf{h}_i(v)$  & Final node representation of $v$ in $G_i$\\ \hline 
		
		$\mathbf{H}^{(l)}_i(V_j)$  &    $l$-th layer representation of nodes $V_j$  in $G_i$\\ \hline 
		
		$\mathbf{H}_i(V_j)$  &  Final representation of nodes $V_j$  in $G_i$\\ \hline 
		
		
		$\mathbf{h}_i$  & Graph representation of  $G_i$\\ \hline 
		
		${y}^*_i$ &  Prediction of GNN $f_\theta$ for  $G_i$ \\  \hline
		
		$\mathbf{S}_i^{(l)}$  & Cluster assignment matrix at $l$-th layer \\ \hline
		
		$\hat{G}^*_i( V_i,\hat{\mathbf{A}}^*_i, \hat{\mathbf{X}}^*_i)$ &  The reconstructed graph for $G_i$  \\ \hline
		
		$\tilde{G}_i,  \hat{G}_i$ &  Augmented positive views for $G_i$  \\ \hline
		
		$s({\mathbf{h}}_i,{\mathbf{h}_j})$ &  Similarity score between $G_i$ and $G_j$  \\ \hline

		$\mathcal{L}_{task}(\cdot)$ & Task loss (Equation~\eqref{eq:task_loss}) \\ \hline
		
		$\mathcal{L}_{ssl}(\cdot)$ & Self-supervised learning loss (Equation~\eqref{eq:ssl_loss}) \\ \hline
		
		$\mathcal{L}_{cl}(\cdot)$ & Contrastive learning loss (Equation~\eqref{eq:cl_loss}) \\
		
		\bottomrule
	\end{tabular}
	
\end{table} 
\begin{table*}[htbp]
	\setlength\tabcolsep{3.9pt}
	\small
	\caption{
		Summary of existing surveys and experimental studies on GNNs for graph-level tasks. Sur. and Exp. denote Survey and Experiments, respectively. 
		Taxo. denotes Taxonomy, Subg. denotes Subgraph. 
		Additionally, FewS., Imba., Effect., and Effic. denote Few-shot, Imbalanced, Effectiveness, and Efficiency, respectively.
	}
	
	\begin{tabular}{c|cc|c|ccccc|cccc|cccc|cc}
		\toprule
		\multirow{2}{*}{\textbf{Paper}} & \multicolumn{2}{c|}{\textbf{Paper Type}}                   & \multirow{2}{*}{\textbf{Taxo.}} & \multicolumn{5}{c|}{\textbf{GNN Type}}                                                                                                                                & \multicolumn{4}{c|}{\textbf{Data Type}}                                                                                & \multicolumn{4}{c|}{\textbf{Scenarios}}                                                                                                  & \multicolumn{2}{c}{\textbf{Eval Metric}}                         \\ \cline{2-3} \cline{5-19} 
		& \multicolumn{1}{c|}{\textbf{Sur.}} & \textbf{Exp.} &                                    & \multicolumn{1}{c|}{\textbf{Node}} & \multicolumn{1}{c|}{\textbf{Pool}} & \multicolumn{1}{c|}{\textbf{Subg.}} & \multicolumn{1}{c|}{\textbf{GL}} & \textbf{SSL} & \multicolumn{1}{c|}{\textbf{SN}} & \multicolumn{1}{c|}{\textbf{BIO}} & \multicolumn{1}{c|}{\textbf{CHE}} & \textbf{MC} & \multicolumn{1}{c|}{\textbf{Clean}} & \multicolumn{1}{c|}{\textbf{Noise}} & \multicolumn{1}{c|}{\textbf{FewS.}} & \textbf{Imba.} & \multicolumn{1}{c|}{\textbf{Effect.}} & \textbf{Effic.} \\ \midrule
		
		\textbf{GNNS}~\cite{wu2020comprehensive}	& \multicolumn{1}{c|}{\ding{51}}                &                     &                                     & \multicolumn{1}{c|}{}              & \multicolumn{1}{c|}{}                 & \multicolumn{1}{c|}{}                  & \multicolumn{1}{c|}{}            &              & \multicolumn{1}{c|}{}            & \multicolumn{1}{c|}{}             & \multicolumn{1}{c|}{}             &             & \multicolumn{1}{c|}{}               & \multicolumn{1}{c|}{}               & \multicolumn{1}{c|}{}                  &                     & \multicolumn{1}{c|}{}                      &                     \\  
		
		\textbf{TUD}~\cite{morris2020tudataset}	& \multicolumn{1}{c|}{}                &            \ding{51}         &                                     & \multicolumn{1}{c|}{\ding{51}}              & \multicolumn{1}{c|}{}                 & \multicolumn{1}{c|}{}                  & \multicolumn{1}{c|}{}            &              & \multicolumn{1}{c|}{\ding{51}}            & \multicolumn{1}{c|}{\ding{51}}             & \multicolumn{1}{c|}{\ding{51}}             &             & \multicolumn{1}{c|}{\ding{51}}               & \multicolumn{1}{c|}{}               & \multicolumn{1}{c|}{}                  &                     & \multicolumn{1}{c|}{\ding{51}}                      &                     \\  
		
		\textbf{OGB}~\cite{hu2020open}	& \multicolumn{1}{c|}{}                &            \ding{51}         &                                     & \multicolumn{1}{c|}{\ding{51}}              & \multicolumn{1}{c|}{}                 & \multicolumn{1}{c|}{}                  & \multicolumn{1}{c|}{}            &              & \multicolumn{1}{c|}{}            & \multicolumn{1}{c|}{\ding{51}}             & \multicolumn{1}{c|}{\ding{51}}             &             & \multicolumn{1}{c|}{\ding{51}}               & \multicolumn{1}{c|}{}               & \multicolumn{1}{c|}{}                  &                     & \multicolumn{1}{c|}{\ding{51}}                      &                     \\  
		
		\textbf{GNNB}~\cite{dwivedi2023benchmarking}	& \multicolumn{1}{c|}{}                &            \ding{51}         &                                     & \multicolumn{1}{c|}{\ding{51}}              & \multicolumn{1}{c|}{}                 & \multicolumn{1}{c|}{\ding{51}}                  & \multicolumn{1}{c|}{}            &              & \multicolumn{1}{c|}{}            & \multicolumn{1}{c|}{\ding{51}}             & \multicolumn{1}{c|}{\ding{51}}             &      \ding{51}       & \multicolumn{1}{c|}{\ding{51}}               & \multicolumn{1}{c|}{}               & \multicolumn{1}{c|}{}                  &                     & \multicolumn{1}{c|}{\ding{51}}                      &                     \\  
		
		\textbf{ReGCB}~\cite{li2024rethinking}		& \multicolumn{1}{c|}{}                &            \ding{51}         &                                    & \multicolumn{1}{c|}{\ding{51}}              & \multicolumn{1}{c|}{}                 & \multicolumn{1}{c|}{}                  & \multicolumn{1}{c|}{}            &              & \multicolumn{1}{c|}{\ding{51}}            & \multicolumn{1}{c|}{\ding{51}}             & \multicolumn{1}{c|}{\ding{51}}             &              & \multicolumn{1}{c|}{\ding{51}}               & \multicolumn{1}{c|}{}               & \multicolumn{1}{c|}{}                  &                     & \multicolumn{1}{c|}{\ding{51}}                      &                     \\  
		
		\textbf{FGNNB}~\cite{errica2020fair}	& \multicolumn{1}{c|}{}                &            \ding{51}         &                                     & \multicolumn{1}{c|}{\ding{51}}              & \multicolumn{1}{c|}{\ding{51}}                 & \multicolumn{1}{c|}{ }                  & \multicolumn{1}{c|}{}            &              & \multicolumn{1}{c|}{\ding{51}}            & \multicolumn{1}{c|}{\ding{51}}             & \multicolumn{1}{c|}{\ding{51}}             &             & \multicolumn{1}{c|}{\ding{51}}               & \multicolumn{1}{c|}{}               & \multicolumn{1}{c|}{}                  &                     & \multicolumn{1}{c|}{\ding{51}}                      &                     \\  
		
		\textbf{GPB}~\cite{wang2024comprehensive}	& \multicolumn{1}{c|}{}                &          \ding{51}           &                                    & \multicolumn{1}{c|}{}              & \multicolumn{1}{c|}{\ding{51}}                 & \multicolumn{1}{c|}{}                  & \multicolumn{1}{c|}{}            &              & \multicolumn{1}{c|}{}            & \multicolumn{1}{c|}{}             & \multicolumn{1}{c|}{}             &             & \multicolumn{1}{c|}{\ding{51}}               & \multicolumn{1}{c|}{\ding{51}}               & \multicolumn{1}{c|}{}                  &                     & \multicolumn{1}{c|}{\ding{51}}                      &                     \\ \midrule
		
		\textbf{OpenGLT}	& \multicolumn{1}{c|}{\ding{51}                  }                & \ding{51}                   &        \ding{51}                                              & \multicolumn{1}{c|}{\ding{51}}              & \multicolumn{1}{c|}{\ding{51}}                 & \multicolumn{1}{c|}{\ding{51}}                  & \multicolumn{1}{c|}{\ding{51}}            &       \ding{51}       & \multicolumn{1}{c|}{\ding{51}}            & \multicolumn{1}{c|}{\ding{51}}             & \multicolumn{1}{c|}{\ding{51}}             &      \ding{51}       & \multicolumn{1}{c|}{\ding{51}}               & \multicolumn{1}{c|}{\ding{51}}               & \multicolumn{1}{c|}{\ding{51}}                  &         \ding{51}            & \multicolumn{1}{c|}{\ding{51}}                      &\ding{51}                    \\ \bottomrule
	\end{tabular}
	\label{tab:related_work}
\end{table*}
\section{GNN for Graph-level Tasks}\label{sec:method}
Recently, GNNs~\cite{DBLP:journals/pvldb/DemirciHF22,zou2023embedx,shao2022decoupled,guliyev2024d3,li2023early,huang2024freshgnn,wang2024tiger,gao2024simple,zhang2023ducati} have emerged as powerful tools for learning node representations by capturing complex relationships within graph structures. Existing GNNs for graph-level tasks can generally be categorized into five types: node-based, pooling-based, subgraph-based, graph learning-based, and self-supervised learning-based GNNs.

\subsection{Node-based GNNs}
As shown in Figure~\ref{fig:GNN4graph}~(a), node-based GNNs learn latent node representations and aggregate them into graph-level representations. Each GNN $f_\theta$ consists of two fundamental operations~\cite{hamilton2017inductive,li2021cache}: $\mathsf{AGG}(\cdot)$ and $\mathsf{COM}(\cdot)$, parameterized by learnable matrices $\mathbf{W}_{agg}$ and $\mathbf{W}_{com}$, respectively. Given a graph $G_i(V_i, \mathbf{A}_i, \mathbf{X}_i)$ and a node $v \in V_i$, the $l$-th layer computes:
\begin{align}
	\mathbf{m}_i^{(l)}(v) &= \mathsf{AGG}^{(l)} \left( \{ (\mathbf{h}_i^{(l-1)}(u), \mathbf{e}_{uv}) : u \in \mathcal{N}_i(v) \} \right) \label{eq:agg}\\
	\mathbf{h}_i^{(l)}(v) &=\sigma \left( \mathsf{COM}^{(l)} \left( \mathbf{h}_i^{(l-1)}(v),  \mathbf{m}_i^{(l)}(v) \right) \right), \label{eq:com}
\end{align}
where $\sigma$ denotes a non-linear function (e.g., ReLU~\cite{li2017convergence}) and $\mathbf{h}_i^{(0)}(v)$ is initialized as $\mathbf{X}_i[v]$. After $L$ layers, we obtain node representations $\mathbf{H}_i(V_i) \in \mathbb{R}^{|V_i| \times d_H}$. The graph representation $\mathbf{h}_i$ is then computed via a permutation-invariant $\mathsf{READOUT}$ function~\cite{li20242} (e.g., $\mathsf{SUM}$, $\mathsf{AVERAGE}$, or $\mathsf{MAX}$):
\begin{align}
	\mathbf{h}_i=\mathsf{READOUT}(\mathbf{H}_i(V_i)).
\end{align}

\noindent {\textbf{Model Optimization.}}
A decoder (e.g., 2-layer MLPs~\cite{zhu2021graph}) predicts a discrete class or continuous property for each graph $G_i$ based on $\mathbf{h}_i$. Given labeled training data $\mathcal{LG}=\{(G_i,y_i)\}_{i=1}^n$, the GNN $f_\theta$ is optimized by:
\begin{align}\label{eq:task_loss}
	& \theta^*= \arg\min_{\theta}\frac{1}{|\mathcal{LG}|}\sum_{G_i \in \mathcal{LG}} {\mathcal{L}_{task}(f_\theta,G_i,y_i)}. \\
	s.t. \quad
	&	\mathcal{L}_{task}(y^*_i,y_i)=
	\begin{cases}  
		- \sum_{y \in {y}_i}y\log y^*_i[y],  &  {y}_i \in \{0,1\}^{|\mathcal{Y}|} \\  
		\Vert y^*_i - y_i\Vert_2, & y_i \in \mathbb{R}
	\end{cases}  
\end{align}

Methods like GCN~\cite{kipf2016semi} and GAT~\cite{velivckovic2017graph} employ tailored aggregation functions, while SATs~\cite{he2022not} can further refine this by excluding irrelevant neighbors to improve representation quality. 
Sampling-based approaches such as GraphSAINT~\cite{zeng2019graphsaint} and GraphSAGE~\cite{hamilton2017inductive}, along with feature-oriented optimization frameworks like SCARA~\cite{liao2022scara}, significantly enhance training scalability. 

Graph Transformers (GTs) extend node-based GNNs by leveraging global attention mechanisms. Representative GTs include Graphormer~\cite{ying2021transformers}, SAN~\cite{kreuzer2021san}, and the Graph Transformer~\cite{dwivedi2020generalization}. 
To improve scalability, GraphGPS~\cite{rampavsek2022recipe} integrates local message passing with global attention, NAGphormer~\cite{chennagphormer} employs efficient neighborhood aggregation, and 
HubGT~\cite{liao2025hubgt} exploits hub labeling to build a label graph with a hierarchical index, completely decoupling graph computation from GT training.

\subsection{Hierarchical Pooling-based GNNs}

\begin{figure*}[t]
	\centering	
	\includegraphics[width = 1\textwidth]{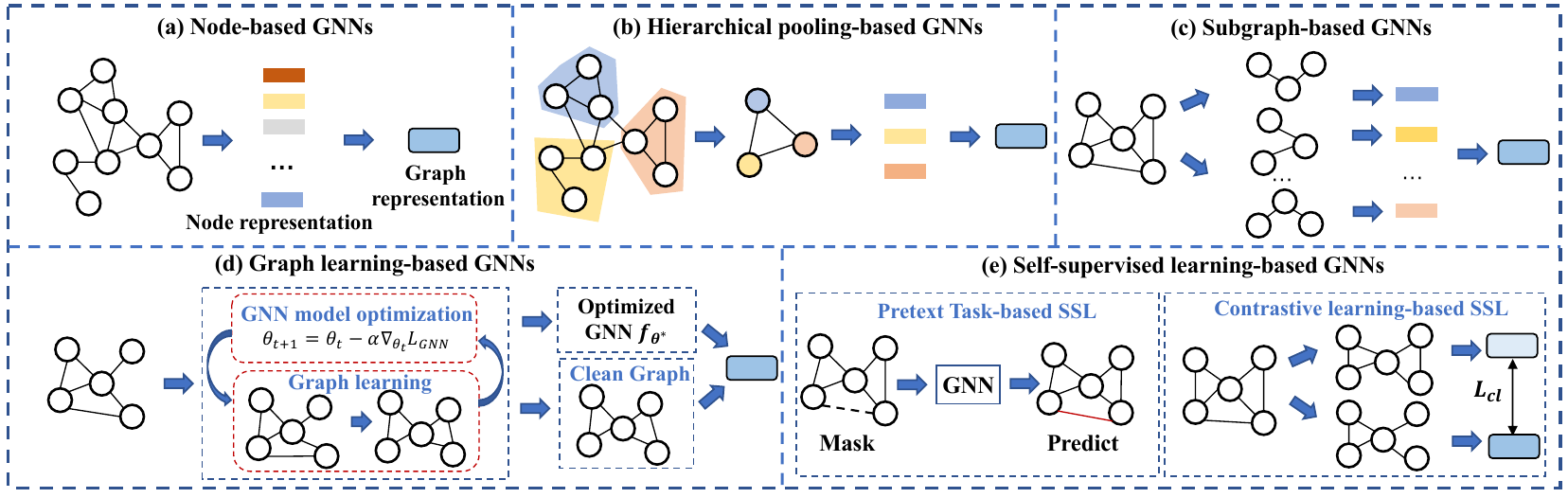}
	\caption{The five types of current GNNs for graph-level tasks.}
	\label{fig:GNN4graph}
\end{figure*}

As shown in Figure~\ref{fig:GNN4graph}~(b), hierarchical pooling (HP)-based 
GNNs progressively coarsen the graph to capture its hierarchical structure 
while preserving essential structural information. At each level, nodes are 
grouped into clusters whose features are summarized, yielding increasingly 
condensed graphs from which the final graph-level representation is derived.

Recall that node-based GNNs compute representations at the $l$-th layer 
using the adjacency matrix $\mathbf{A}_i$ and hidden representations 
$\mathbf{H}_i^{(l-1)}(V_i)$ via Equations~\eqref{eq:agg} 
and~\eqref{eq:com}. HP-based GNNs instead first produce a cluster 
assignment matrix 
$\mathbf{S}_i^{(l)} \in \{0,1\}^{n_{l-1} \times n_l}$ with $n_l < n_{l-1}$, 
which maps each node in $V_i^{(l-1)}$ to one of $n_l$ clusters. 
A coarsened adjacency matrix 
$\mathbf{A}_i^{(l)} \in \mathbb{R}^{n_l \times n_l}$ is then obtained as:
\begin{align}
	\mathbf{A}_i^{(l)} = \mathbf{S}_i^{(l)\top} \mathbf{A}_i^{(l-1)} \mathbf{S}_i^{(l)}.
\end{align}
The $n_l$ resulting clusters become the new node set $V_i^{(l)}$, and their 
initial hidden representations are computed by applying a \textsf{READOUT} 
operation over the members of each cluster $k \in \{1, \ldots, n_l\}$:
\begin{align}
	\mathbf{H}^{(l-1)}_i(V^{(l)}_i)[{k}] = \textsf{READOUT}(\ \mathbf{H}^{(l-1)}_i(V^{(l-1)}_i)[V_{k}]),
\end{align}
where $V_k = \{u \mid \mathbf{S}_i^{(l)}[u][k] = 1\}$ denotes the set of 
nodes assigned to cluster $k$.

Depending on how the assignment matrix $\mathbf{S}_i^{(l)}$ is constructed, 
existing HP-based GNNs can be categorized into three types.

\begin{itemize}[leftmargin=10pt]

	\item \textbf{Similarity-based.}
	These methods~\cite{mesquita2020rethinking,dhillon2007weighted,li2022cc} cluster nodes using predefined similarity metrics (e.g., cosine similarity of features) or graph partitioning algorithms. For instance, Graclus~\cite{mesquita2020rethinking,dhillon2007weighted} and CC-GNN~\cite{li2022cc} assign nodes to clusters based on feature similarity and graph structure.
	
	\item \textbf{Node Dropping-based.}
	These methods~\cite{cangea2018towards,gao2019graph,lee2019self} learn an importance score for each node and retain only the top-$n_l$ nodes at layer $l$, effectively assigning one node per cluster and dropping the rest. Representative methods include TopKPool~\cite{cangea2018towards,gao2019graph} and SAGPool~\cite{lee2019self}.
	
	\item \textbf{Learning-based.}
	These methods~\cite{ying2018hierarchical,bianchi2020spectral,vaswani2017attention,DBLP:conf/iclr/BaekKH21,diehl2019edge} use neural networks to learn the cluster assignment matrix $\mathbf{S}_i^{(l)}$ from node features and graph structure. DiffPool~\cite{ying2018hierarchical} and MinCutPool~\cite{bianchi2020spectral} employ non-linear networks, GMT~\cite{DBLP:conf/iclr/BaekKH21} leverages multi-head attention~\cite{vaswani2017attention}, and EdgePool~\cite{diehl2019edge} learns edge scores between connected nodes to construct the cluster matrix.
\end{itemize}

\subsection{Subgraph-based GNNs}
Recent studies introduce subgraph-based GNNs that achieve stronger expressive 
power by explicitly capturing substructure information. 
As shown in Figure~\ref{fig:GNN4graph}~(c), these methods decompose an input 
graph into a collection of (possibly overlapping) subgraphs and learn 
representations for each one to enrich the final graph-level embedding.
Formally, given a graph $G_i(V_i, \mathbf{A}_i, \mathbf{X}_i)$, a set of 
$n_s$ subgraphs $\{G_{i,j}(V_{i,j}, \mathbf{A}_{i,j}, \mathbf{X}_{i,j})\}_{j=1}^{n_s}$ 
is first extracted. Node representations within each subgraph $G_{i,j}$ are 
then computed via Equations~\eqref{eq:agg} and~\eqref{eq:com}. 
Since a node $v \in V_i$ may belong to multiple subgraphs, it can receive 
multiple representations, which are merged into a single embedding 
$\mathbf{h}_i(v)$ through a \textsf{READOUT} function:
\begin{align}
	\mathbf{h}_i(v) = \textsf{READOUT}(\mathbf{h}_{i,j}(v)\mid v \in V_{i,j}).
\end{align}

Existing subgraph-based methods can be categorized into three types according 
to how the subgraphs are constructed.
\begin{itemize}[leftmargin=10pt]

	\item \textbf{Graph Element Deletion-based.}
	These approaches~\cite{cotta2021reconstruction,ding2024sgood,papp2021dropgnn,bevilacqua2022equivariant} delete specific nodes or edges to create subgraphs, enabling GNNs to focus on the most informative parts. For example, DropGNN~\cite{papp2021dropgnn} and ESAN~\cite{bevilacqua2022equivariant} generate subgraphs via random edge deletion to enhance expressiveness, while SGOOD~\cite{ding2024sgood} abstracts a superstructure from original graphs and applies sampling and edge deletion on it to create more diverse subgraphs.
	
	\item \textbf{Rooted Subgraph-based.}
	These approaches~\cite{bevilacqua2024efficient,you2021identity,yan2024efficient,frasca2022understanding,huang2022boosting,papp2022theoretical,qian2022ordered,yang2023extract,zhang2021nested,zhao2022from} generate subgraphs centered around specific \textit{root nodes} to capture their structural roles and local topology, thereby enhancing GNN expressiveness. I2GNN~\cite{huang2022boosting}, ECS~\cite{yan2024efficient}, and ID-GNN~\cite{you2021identity} append positional side information to root nodes, such as ID identifiers~\cite{huang2022boosting,you2021identity} or node degree and shortest distance~\cite{yan2024efficient}. NestGNN~\cite{zhang2021nested} and GNN-AK~\cite{zhao2022from} use rooted subgraphs with varying hops to capture hierarchical relationships.
	
	\item \textbf{$k$-hop Subgraph-based.}
	These approaches~\cite{nikolentzos2020k,feng2022powerful,abu2019mixhop,yao2023improving,sandfelder2021ego,wang2020multi,wang2024tiger,ye2025mose} construct subgraphs based on the $k$-hop neighborhood of each node, aggregating information not only from 1-hop neighbors but also directly from nodes up to $k$ hops away. MixHop~\cite{abu2019mixhop} uses a graph diffusion kernel to gather multi-hop neighbors. SEK-GNN~\cite{yao2023improving}, KP-GNN~\cite{feng2022powerful}, EGO-GNN~\cite{sandfelder2021ego}, and $k$-hop GNN~\cite{nikolentzos2020k} progressively update node representations by aggregating within $k$-hops. MAGNA~\cite{wang2020multi} learns pairwise node weights based on all paths within $k$-hops.
\end{itemize}

\subsection{Graph Learning-based GNNs}
Due to uncertainty and complexity in data collection, real-world graphs often contain redundant, biased, or noisy edges and features. 
When operating on such imperfect structures, vanilla GNNs may learn spurious correlations and thus fail to produce reliable graph representations, ultimately leading to incorrect predictions. 
To mitigate this issue, as shown in Figure~\ref{fig:GNN4graph}~(d), 
recent work~\cite{fatemi2023ugsl,zhiyao2024opengsl,li2024gslb} propose to learn from purified or reconstructed graph structure and enhanced node features that better reflect the underlying signal to improve the quality of the learned graph representations. 
Given a labeled graph set $\mathcal{LG}=\{(G_i,y_i)\}_{i=1}^n$, the graph learning-based approaches can be formulated as bi-level optimization problem.
\begin{align}
	\label{eq:GL_high}
	&	\theta^* = \min_{\theta \in \Theta}{\frac{1}{|\mathcal{LG}|}\sum_{G_i \in \mathcal{LG}}\mathcal{L}_{task}({f}_\theta, \hat{G}^*_i( V_i,\hat{\mathbf{A}}^*_i, \hat{\mathbf{X}}^*_i), y_i)}. \\
	s.t. \quad 
	&\hat{\mathbf{A}}^*_i, \hat{\mathbf{X}}^*_i = \arg\min_{\hat{\mathbf{A}}_i, \hat{\mathbf{X}}_i} \mathcal{L}_{gl}\bigl(f_{\theta^*}, \hat{G}_i( V_i, \hat{\mathbf{A}}_i, \hat{\mathbf{X}}_i), y_i\bigr),\notag\\
	&\hfill \forall G_i \in \mathcal{G}\label{eq:GL_low}
\end{align}
At the low level in Equation~\eqref{eq:GL_low}, current approaches propose different graph learning objectives $\mathcal{L}_{gl}(\cdot)$  to reconstruct  graph structure $\mathbf{A}_i$ and node features $\mathbf{X}_i$. 
Then, in Equation~\eqref{eq:GL_high}, $\hat{G}^*_i( V_i, \hat{\mathbf{A}}^*_i, \hat{\mathbf{X}}^*_i)$  will be used to optimize the GNNs by the loss function in Equation~\eqref{eq:task_loss}. 

Depending on the techniques of reconstructing graphs, current GL-based GNNs can be categorized into three types.
\begin{itemize}[leftmargin=10pt]
	\item \textbf{Preprocessing-based.}
	These approaches~\cite{wu2019adversarial,li2022black,entezari2020all} reconstruct graphs before training by recovering common graph patterns. GNN-Jaccard~\cite{wu2019adversarial} and GNAT~\cite{li2022black} remove edges between dissimilar nodes and add edges between similar ones, based on the homophily assumption. GNN-SVD~\cite{entezari2020all} reconstructs graphs by reducing the rank of the adjacency matrix, as noisy edges tend to increase it.
	
	\item \textbf{Jointly Training-based.}
	Unlike static preprocessing, these approaches~\cite{jin2021node,jin2020graph,li2024fight,franceschi2019learning,sun2022graph,luo2021learning,zhang2019hierarchical,zhou2024motif,wang2023prose} iteratively reconstruct the graph structure and node features alongside GNN optimization through bi-level optimization. ADGNN~\cite{li2024fight}, ProGNN~\cite{jin2020graph}, and SimPGCN~\cite{jin2021node} reconstruct edges by jointly minimizing the GNN loss and the rank of the adjacency matrix. Alternatively, MOSGSL~\cite{zhou2024motif} and HGP-SL~\cite{zhang2019hierarchical} first partition graphs into subgraphs based on node similarities and predefined motifs, then reconstruct edges at the subgraph level rather than the node level.
\end{itemize}

\subsection{Self Supervised Learning-based GNNs}
Self-supervised learning (SSL) has become a powerful paradigm to pretrain GNNs without the need for labeled data, which can capture the node patterns and graph patterns.
As shown in Figure~\ref{fig:GNN4graph}~(e),
the key idea of SSL approaches is to create supervised signals directly from the structure and node features of the unlabeled graph itself, leveraging the graph's inherent properties to guide the learning process.
Formally, given a set of unlabeled graphs $\mathcal{UG}=\{G_i(V_i, \mathbf{A}_i, \mathbf{X}_i)\}_{i=1}^{|\mathcal{UG}|}$, the GNN $f_\theta$  is pretrained as follows:
\begin{align}
	\label{eq:ssl_loss}
	\theta^\prime = \arg\min_{\theta}\frac{1}{|\mathcal{UG}|}\sum_{G_i \in \mathcal{UG}}\mathcal{L}_{ssl}(f_\theta, G_i, Signal_i ),
\end{align}
where $Signal_i$ is the supervised signals from the unlabeled graph $G_i$ and $\theta^\prime$ is the optimized GNN parameters.
Then, the pretrained $f_{\theta^\prime}$ can be used to predict graph labels or properties. Formally, given the set of labeled graphs $\mathcal{LG}=\{G_j(V_j \mathbf{A}_j, \mathbf{X}_j), y_j\}_{j=1}^{|\mathcal{LG}|}$, the GNN $f_{\theta^\prime}$  is optimized as follows:
\begin{align}
	\theta^* = \arg\min_{\theta^\prime}\frac{1}{|\mathcal{LG}|}\sum_{G_j \in \mathcal{LG}}\mathcal{L}_{task}(f_{\theta^\prime}, G_j, y_j ),
\end{align}
where the task loss $\mathcal{L}_{task}(\cdot)$ is defined in Equation~\eqref{eq:task_loss}.

Depending on the specific technique used to auto-generate supervised signals from unlabeled graphs, SSL-based GNN approaches can be broadly categorized into two main paradigms.

\begin{itemize}[leftmargin=10pt]
	\item \textbf{Pretext Task-based.}
	These approaches~\cite{hou2022graphmae,inae2023motif,zhang2021motif,zang2023hierarchical,hu2020gpt,jin2020self,wang2022graph} design auxiliary tasks to learn representations from graph structure and features without external labels, such as predicting node attributes, node degrees, or node counts~\cite{wang2022graph}. For example, HMGNN~\cite{zang2023hierarchical} predicts links and node counts; MGSSL~\cite{zhang2021motif} masks and predicts edges among motifs; MoAMa~\cite{inae2023motif} masks and reconstructs node features; GraphMAE~\cite{hou2022graphmae} and GPTGNN~\cite{hu2020gpt} predict both node attributes and edges.
	
	\item \textbf{Graph Contrastive Learning-based.}
	Graph contrastive learning (GCL)-based approaches~\cite{wang2022augmentation,perozzi2014deepwalk,lee2022augmentation,hassani2020contrastive,yuan2021semi,sun2024motif} learn representations by maximizing the similarity between augmented views of the same graph (positive pairs) while minimizing similarity with different graphs (negative pairs).
	The SSL loss $\mathcal{L}_{ssl}(\cdot)$ in Equation~\eqref{eq:ssl_loss} can be formulated as:
	\begin{align}\label{eq:cl}
		&	\theta^\prime = \arg\min_{\theta}{\frac{1}{|\mathcal{UG}|}\sum_{G_i \in \mathcal{UG}}\mathcal{L}_{cl}(f_\theta, \hat{G}_i, \tilde{G}_i, Neg_i)}. \\
		s.t.\ & \tilde{G}_i,  \hat{G}_i = \arg\min_{\tilde{G}_i,  \hat{G}_i}{\mathcal{L}_{positive}(G_i,\mathcal{T})},\forall G_i \in \mathcal{UG}, \label{eq:postive_view}
	\end{align}
	where $\mathcal{L}_{cl}(\cdot)$ is the contrastive loss, $\mathcal{L}_{positive}(\cdot)$ generates two positive views ($\tilde{G}_i$ and $\hat{G}_i$) using augmentation operations $\mathcal{T}$, and $Neg_i$ are negative samples typically drawn from other graphs.
	A typical contrastive loss based on InfoNCE~\cite{zhu2021graph,yeh2022decoupled} is:
	\begin{align}\label{eq:cl_loss}
		\mathcal{L}_{cl}(\cdot) = - \log \frac{\text{s}(\mathbf{\hat{h}}_i, \mathbf{\tilde{h}}_i) }{\sum_{G^\prime_i \in \{\hat{G}_i,\tilde{G}_i\}}\sum_{G_j \in A(G_i) \}}s(\mathbf{\hat{h}}^\prime_i, \mathbf{\tilde{h}}_j) 
		}, 
	\end{align}
	where $A(G_i)= \hat{G}_i \cup \tilde{G}_i \cup {Neg}_i$, $\hat{\mathbf{h}}_i$ and $\tilde{\mathbf{h}}_i$ are the representations of $\hat{G}_i$ and $\tilde{G}_i$, and $s({\mathbf{h}}_i,{\mathbf{h}_j})=\exp(\text{cosine}({\mathbf{h}}_i,{\mathbf{h}}_j)/\tau)$ is a temperature-scaled similarity score.
	
	\quad 
For generating positive views (Equation~\eqref{eq:postive_view}), 
similarity-based methods~\cite{wang2022augmentation,perozzi2014deepwalk,lee2022augmentation} 
pair structurally or feature-wise similar nodes, 
diffusion-based methods~\cite{hassani2020contrastive,yuan2021semi,sun2024motif} 
reshape topology via global propagation such as personalized 
PageRank~\cite{haveliwala1999efficient} or motif-preserving 
diffusion~\cite{sun2024motif}, and perturbation-based 
methods~\cite{thakoor2021bootstrapped,zhu2020graph,zhu2021graph,li20242} 
stochastically modify edges and node attributes. 
View quality can be further improved through learnable or adversarial 
generators~\cite{pu2023graph,suresh2021adversarial}, automated augmentation 
search~\cite{luoautomated,you2021graph}, robust 
perturbations~\cite{kong2022robust}, invariance-driven 
regularization~\cite{liu2022graph,wu2022dir,yuan2021semi}, and hybrid 
generative-contrastive frameworks~\cite{wang2024generative}.
\end{itemize}

\section{Benchmark Design}\label{sec:benchmark_design}

\subsection{Evaluation Framework OpenGLT} \label{ssec:opengtl}
\begin{figure}[t]
	\centering	
	\includegraphics[width = 0.47\textwidth]{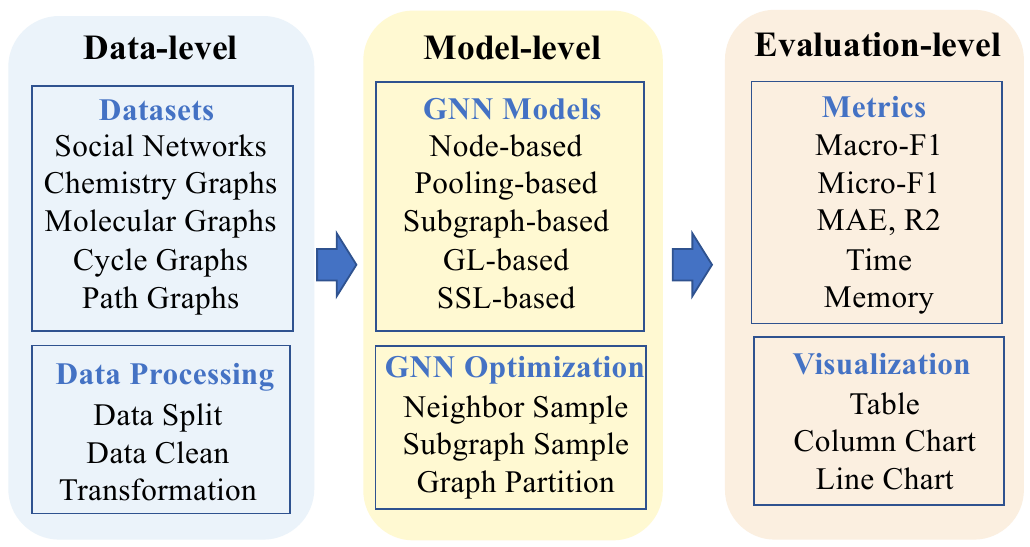}
	\caption{Evaluation framework. }
	\label{fig:pipeline}
\end{figure}

OpenGLT is built on five principles:
\textbf{(P1)~Principled Coverage}: datasets and models are selected
to systematically span the graph-property space and architectural
design space;
\textbf{(P2)~Fairness}: all models share identical splits,
tuning budgets, and hardware;
\textbf{(P3)~Comprehensiveness}: evaluation covers diverse domains,
task types, and realistic scenarios;
\textbf{(P4)~Reproducibility}: all code and configs are
publicly released;
\textbf{(P5)~Extensibility}: a modular design supports seamless
addition of new models, datasets, and metrics.
As shown in Figure~\ref{fig:pipeline}, the framework comprises three
levels.  The \emph{data level} manages datasets across four domains
with unified preprocessing, splitting, and scenario construction
(noise, imbalance, few-shot).  The \emph{model level} wraps 20 GNNs
from all five categories (Section~\ref{sec:gnn_graph}) in a common
training interface with scalable optimization.  The
\emph{evaluation level} computes effectiveness metrics (Accuracy,
Micro/Macro-F1, MAE, \(R^2\)) and efficiency metrics (time, memory),
with automated visualization.

%
%
%

\subsection{Datasets}\label{ssec:dataset}
%
%
We evaluate GNNs on four domains {\color{black} across 26 datasets}: social networks (SN), biology (BIO), chemistry (CHE), and motif counting (MC).  For dataset splitting, we use widely adopted standard splits when available, or otherwise 10-fold cross-validation. 
Detailed statistics are in Table~\ref{tab:data}, respectively. 

\begin{itemize}[leftmargin=10pt]
	\item \textbf{Social Networks:} We use \textbf{IMDB-BINARY}~\cite{morris2020tudataset} and \textbf{IMDB-MULTI}~\cite{morris2020tudataset} for movie genre classification, \textbf{REDDIT-BINARY}~\cite{morris2020tudataset} for discussion thread classification, and \textbf{COLLAB}~\cite{morris2020tudataset} for research field classification in co-authorship networks.
	
	\item \textbf{Biology:} We utilize five datasets in the biological domain. Protein datasets include \textbf{PROTEINS}~\cite{morris2020tudataset} and \textbf{DD}~\cite{morris2020tudataset} whose task is binary classification on distinguishing between enzymes and non-enzymes, and \textbf{ENZYMES}~\cite{morris2020tudataset} for multi-class classification assigning proteins to one of the six top-level Enzyme Commission (EC) classes. 
	Molecular property prediction datasets include \textbf{MolHIV}~\cite{hu2020open} performing binary classification to predict whether a molecule inhibits HIV replication, while \textbf{MolTox21}~\cite{hu2020open} is a multi-label classification task predicting the presence of toxicity across 12 different assays.
	
	\item \textbf{Chemistry:} We use four molecular graph datasets where nodes represent atoms and edges denote chemical bonds. 
	\textbf{MUTAG}~\cite{morris2020tudataset} and \textbf{NCI1}~\cite{morris2020tudataset} are binary classification tasks predicting the mutagenicity of compounds and the anti-cancer activity, respectively. 
	\textbf{MolBACE}~\cite{hu2020open} is designed to predict inhibitors of BACE-1, a crucial enzyme in Alzheimer’s disease. 
	And \textbf{MolPCBA}~\cite{hu2020open} is a multi-label classification dataset containing 128 bioassays.
	
	\item \textbf{Motif Counting:} We employ synthetic datasets with task of predicting the exact occurrences of 13 different specific subgraph structures (motifs) within a given graph, such as cycles and paths. 
	These datasets serves as a benchmark for evaluating a GNN’s expressiveness in capturing and reasoning over local structural patterns, which are sourced from~\cite{chen2020can} and are widely used in recent works~\cite{chen2020can,zhao2022from,huang2022boosting,yan2024efficient}.
\end{itemize}

\begin{table}[t]
	\vspace{-0.5em}
	\caption{Data statistics. \textoverline{Nodes} and \textoverline{Edges} denote the average number of nodes and edges per graph, respectively.
		Split is the data partition strategy, where 10-fold denotes 10-fold cross-validation, others denote train/validation/test splits.
	}
	
	\label{tab:data}
	\setlength\tabcolsep{3.2pt}
	\small
	\begin{tabular}{c|c|c|c|c|c|c}
		\hline
		\textbf{Type} & \textbf{Dataset} & \textbf{Graphs} & \textbf{$\overline{\textbf{Nodes}}$} 
		& \textbf{$\overline{\textbf{Edges}}$} & \textbf{Classes} & \textbf{Split} \\
		\hline
		\multirow{4}{*}{SN}
		& IMDB-B (I-B)   	& 1,000   & 19.8   & 96.5   & 2  & 10-fold \\
		& IMDB-M (I-M)    	& 1,500   & 13.0   & 65.9   & 3  & 10-fold \\
		& REDDIT-B (RED) 	& 2,000   & 429.6  & 497.8  & 2  & 10-fold \\
		& COLLAB (COL)   	& 5,000   & 74.5   & 2457.8 & 3  & 10-fold \\
		\hline
		\multirow{5}{*}{BIO}
		& PROTEINS (PRO) 	& 1,113   & 39.1   & 72.8   & 2  & 10-fold \\
		& DD       			& 1,178   & 284.3  & 715.7  & 2  & 10-fold \\
		& ENZYMES (ENZ)  	& 600     & 32.6   & 62.1   & 6  & 10-fold \\
		& MolHIV (HIV)  	& 41,127  & 25.5   & 54.9   & 2  & 8/1/1   \\
		& MolTox21 (TOX) 	& 7,831   & 18.9   & 39.2   & 12 & 8/1/1   \\
		\hline
		\multirow{4}{*}{CHE}
		& MUTAG (MUT)   	& 188     & 17.9   & 19.8   & 2  & 10-fold \\
		& NCI1 (NCI)    	& 4,110   & 29.9   & 32.3   & 2  & 10-fold \\
		& MolBACE (BAC) 	& 1,513   & 34.1   & 36.9   & 2  & 8/1/1   \\
		& MolPCBA (PCB) 	& 437,929 & 26.0   & 28.1   & 128& 8/1/1   \\
		\hline
		\multirow{7}{*}{MC}
		& \{3,4,5,6,7,8\}-Cycle	& 5,000 & 18.8 & 31.3 & 1 & 3/2/5 \\
		& \{4,5,6\}-Path        & 5,000 & 18.8 & 31.3 & 1 & 3/2/5 \\
		& 4-Clique              & 5,000 & 18.8 & 31.3 & 1 & 3/2/5 \\
		& Tailed Tri.           & 5,000 & 18.8 & 31.3 & 1 & 3/2/5 \\
		& Chor. Cyc.         & 5,000 & 18.8 & 31.3 & 1 & 3/2/5 \\
		& Tri. Rec.    & 5,000 & 18.8 & 31.3 & 1 & 3/2/5 \\
		\hline
	\end{tabular}
	\vspace{-2em}
\end{table}

\subsection{Evaluated GNNs}\label{ssec:baselines}

We comprehensively evaluate 20 representative and effective GNNs across five categories,  as follows:

\begin{itemize}[leftmargin=10pt]
	
	\item \textbf{Node-based GNNs (7 models).}
	We include classic message-passing networks alongside expressive aggregation 
	strategies and modern Graph Transformers (GTs) to span the full range of 
	node-level design choices.
	(1)~\textbf{\textit{GCN}}~\cite{kipf2016semi} and 
	(2)~\textbf{\textit{GraphSAGE (SAGE)}}~\cite{hamilton2017inductive} 
	learn node representations with neighbor sampling for scalability.
	(3)~\textbf{\textit{GIN}}~\cite{xu2018how} updates nodes using a sum of 
	neighbor features followed by an MLP, achieving discriminative power equivalent 
	to the 1-WL test.
	(4)~\textbf{\textit{PNA}}~\cite{corso2020principal} combines multiple 
	aggregators with degree-based scalers to capture richer neighbor distributions.
	(5)~\textbf{\textit{GraphGPS (GPS)}}~\cite{rampavsek2022recipe} represents the 
	culmination of Graph Transformers by modularly integrating local message passing 
	with global attention.
	As standard GTs are   computationally intensive, we further select two 
	representative acceleration techniques: 
	(6)~\textbf{\textit{NAGphormer (NAG)}}~\cite{chennagphormer}, which employs 
	hop-based neighbor tokenization, and 
	(7)~\textbf{\textit{HubGT (HGT)}}~\cite{liao2025hubgt}, which exploits 
	decoupled hub-based hierarchical indexing.
	
	\item \textbf{HP-based GNNs (3 models).}
	We cover the three main pooling paradigms, including node dropping, learning-based clustering, 
	and edge contraction, to reflect the diversity of hierarchical coarsening.
	(8)~\textbf{\textit{TopK}}~\cite{gao2019graph} selects top-scoring nodes based 
	on trainable projection scores to construct coarser graphs.
	(9)~\textbf{\textit{GMT}}~\cite{DBLP:conf/iclr/BaekKH21} employs 
	transformer-based attention to adaptively group nodes into hierarchical clusters.
	(10)~\textbf{\textit{EdgePool (EP)}}~\cite{diehl2019edge} contracts the most 
	significant edges to merge connected nodes hierarchically.
	
	\item \textbf{Subgraph-based GNNs (4 models).}
	We select models that collectively represent the major subgraph construction 
	strategies, including, structural encoding, $k$-hop aggregation, and root-based 
	identification, together with a recent efficiency-oriented variant.
	(11)~\textbf{\textit{ECS}}~\cite{wang2022neural} integrates structural 
	embeddings and encodes subgraph distances to distinguish substructures for 
	counting tasks.
	(12)~\textbf{\textit{GNNAK+ (AK+)}}~\cite{zhao2022from} aggregates information 
	from $k$-hop subgraphs to capture high-order structures.
	(13)~\textbf{\textit{I2GNN (I2)}}~\cite{huang2022boosting} utilizes unique 
	identifiers for subgraph roots and neighbors to distinguish structural roles.
	(14)~\textbf{\textit{HyMN (HMN)}}~\cite{southernbalancing} represents the 
	latest acceleration technique for resource-intensive subgraph GNNs, significantly 
	reducing computational complexity by selectively processing   subgraphs 
	guided by walk-based centrality.
	
	\item \textbf{GL-based GNNs (3 models).}
	We choose methods that cover complementary graph refinement principles—information-theoretic 
	filtering, attention-based selection, and motif-driven reconstruction—to assess how 
	different structure learning objectives affect downstream performance.
	(15)~\textbf{\textit{VIBGSL (VIB)}}~\cite{sun2022graph} employs the Information 
	Bottleneck principle to learn task-relevant structures.
	(16)~\textbf{\textit{HGP-SL (HGP)}}~\cite{zhang2019hierarchical} selects and 
	refines subgraphs using sparse attention mechanisms.
	(17)~\textbf{\textit{MOSGSL (MO)}}~\cite{zhou2024motif} dynamically reconstructs 
	motif-driven subgraphs to align discriminative patterns.
	
	\item \textbf{SSL-based GNNs (3 models).}
	We include representative contrastive learning methods that differ in their view 
	generation strategies—geometry-aware, diffusion-based, and adaptive 
	perturbation-based—to evaluate how pretraining signals transfer to graph-level tasks.
	(18)~\textbf{\textit{RGC}}~\cite{sun2024motif} uses diverse-curvature GCNs and 
	motif-aware contrastive objectives.
	(19)~\textbf{\textit{MVGRL (MVG)}}~\cite{hassani2020contrastive} contrasts 
	embeddings from original and diffusion-based graph views.
	(20)~\textbf{\textit{GCA}}~\cite{zhu2021graph} contrasts node embeddings across 
	adaptively augmented graph views to capture the shared   features.
	
\end{itemize}

\subsection{Evaluation  Metric}\label{ssec:evaluation}
We evaluate the performance of GNNs using effectiveness and efficiency metrics as follows.

\subsubsection{Effectiveness Metric}
Given a graph set $\mathcal{LG}=\{(G_i, y_i)\}_{i=1}^{|\mathcal{LG}|}$,  we denote the prediction label of each graph $G_i$ as $\hat{y}_i$.
For graph classification tasks, we use the \textbf{Strict Accuracy (Acc)}, \textbf{Micro-F1 (Mi-F1)}, and \textbf{Macro-F1 (Ma-F1)}.
Particularly, if each graph only has one label, 
the \textbf{Micro-F1} is same as \textbf{Accuracy}.
\begin{itemize}[leftmargin=12pt]
	
	\item \textbf{Strict Accuracy (Acc).}
	Strict accuracy is used to measure the proportion of exact matches between predicted and true labels.  
	Strict accuracy is defined as  $Acc = \frac{1}{|\mathcal{LG}|}\sum_{G_i \in \mathcal{LG}}{\mathbb{I}(y_i = \hat{y}_i)}$, where   $\mathbb{I}(y_i = \hat{y}_i) = 1$ if only $y_i = \hat{y}_i$.
	
	\item \textbf{Micro-F1 (Mi-F1).}
	Micro-F1 is a performance metric that considers the overall precision and recall across all instances in the dataset.  
	The Micro-precision is defined as $Mi\text{-}P=\frac{\sum_{G_i \in \mathcal{LG}}{|y_i \cap \hat{y}_i|}}{\sum_{G_j \in \mathcal{LG}}{|\hat{y}_j|} }$ and  Micro-recall is $Mi\text{-}R=\frac{\sum_{G_i \in \mathcal{LG}}{|y_i \cap \hat{y}_i|}}{\sum_{G_j \in \mathcal{LG}}{|{y}_j|} }$.
	Then, the Micro-F1 is defined as $Mi\textit{-}F1=\frac{2 \times Mi\text{-}P \times Mi\text{-}R}{Mi\text{-}P + Mi\text{-}R}$.
	
	\item \textbf{Macro-F1 (Ma-F1).}
	Macro-F1 evaluates the average performance of precision and recall across all instances, treating each equally regardless of size.  
	The Macro-precision is defined as $Ma\text{-}P = \frac{1}{|\mathcal{LG}|} \sum_{G_i \in \mathcal{LG}} \frac{|y_i \cap \hat{y}_i|}{|\hat{y}_i|}$ and Macro-recall is defined  as $Ma\text{-}R = \frac{1}{|\mathcal{LG}|} \sum_{G_i \in \mathcal{LG}} \frac{|y_i \cap \hat{y}_i|}{|y_i|}$, 
	so the Macro-F1 is defined as  $Ma\text{-}F1 = \frac{2 \times Ma\text{-}P \times Ma\text{-}R}{Ma\text{-}P + Ma\text{-}R}$.
	
\end{itemize}

For graph regression tasks, we use the \textbf{Mean Absolute Error (MAE)} and \textbf{R2} as follows.
\begin{itemize}[leftmargin=12pt]
	
	\item \textbf{Mean Absolute Error (MAE).}
	Mean Absolute Error measures the average magnitude of errors between predicted and true values.  
	MAE  is defined as  $MAE = \frac{1}{|\mathcal{LG}|} \sum_{G_i \in \mathcal{LG}} {|y_i - \hat{y}_i|}$.
	Lower $MAE$ indicates better performance. 
	
	\item \textbf{R2.}
	R2 evaluates the proportion of variance in the true values that is captured by the predicted values.  
	The R2  is defined as $R2 = 1 - \frac{\sum_{G_i \in \mathcal{LG}} (y_i - \hat{y}_i)^2}{\sum_{G_i \in \mathcal{LG}} (y_i - \bar{y})^2}\in[0,1]$, where $\bar{y} = \frac{1}{|\mathcal{LG}|} \sum_{G_i \in \mathcal{LG}} y_i$ is  the mean of the true values. 
	Higher $R2$ indicates better performance. 
	
\end{itemize}

\subsubsection{Efficiency Metric}
We evaluate the efficiency of models on both graph classification and regression tasks based on the \textbf{training time (s), inference time (s)},   
\textbf{memory usage (MB)} in training and inference phases.

\subsection{Hyperparameter and Hardware Setting}\label{appx:ssec:hyper}
For classification datasets, we set the batch size as 32 for four larger datasets (REDDIT, COLLAB, DD, and MolPCBA) and 128 for the other datasets. 
For regression datasets, we set the batch size as 256. 
To efficiently tune the hyperparameters for each model, we employed the Optuna framework~\cite{akiba_optuna}. 
For each model, we conducted $200$ trials using the Tree-structured Parzen Estimator (TPE) sampler and a MedianPruner to terminate unpromising trials early. 
The hyperparameter search spaces were defined as follows: 
the hidden dimension is selected from $\{64, 128, 256, 512\}$,
the learning rate from $\{1e-2, 1e-3, 1e-4, 1e-5\}$, 
GNN layers from $\{1,2,3,4\}$,
and the dropout rate from $\{0, 0.1, 0.2, 0.3, 0.4, 0.5\}$. 
All models are trained for a maximum of 2000 epochs, with early stopping applied if no improvement is observed on the validation set within 50 epochs.

All experiments are  executed on a CentOS 7 machine equipped with dual 10-core Intel® Xeon® Silver 4210 CPUs @ 2.20GHz, 8 NVIDIA GeForce RTX 2080 Ti GPUs (11GB each) and 256GB RAM.

\section{Results}\label{sec:experiments}

\subsection{Effectiveness Evaluation}\label{ssec:effectiveness}

\begin{table*}
	\vspace{-1em}
    \caption{
		Evaluation on graph classification. All results are reported as percentages (\%). 
		The best and second-best results are highlighted in bold and underlined, respectively. 
		``\textit{Met.}'' denotes Metrics, ``OOM'' indicates out-of-memory, and ``TLE'' represents that training could not be completed within a time limit of 3 days.
	}
    \centering
    \setlength\tabcolsep{0.55pt}
    \renewcommand{\arraystretch}{1.1}
    \begin{tabular}{|c|c|c|c|c|c|c|c|c|c|c|c|c|c|c|c|c|c|c|c|c|c|c|}
        \hline
        \multirow{2}{*}{\textbf{Type}}      & \multirow{2}{*}{\textbf{Data}}          & \multirow{2}{*}{\textit{\textbf{Met.}}}  & \multicolumn{7}{c|}{\textbf{Node-based}}                                                                                                                                                                           & \multicolumn{3}{c|}{\textbf{Pooling-based}}                                  & \multicolumn{4}{c|}{\textbf{Subgraph-based}}                                                   & \multicolumn{3}{c|}{\textbf{GL-based}}                                           & \multicolumn{3}{c|}{\textbf{SSL-based}}                                          \\ \cline{4-23} 

                                            &                                         &                                     & GCN                   & GIN                   & SAGE                  & PNA                   & NAG                   & HGT                   & GPS                   & TopK                  & GMT                   & EP                    & ECS                   & AK+                   & I2                    & HMN                  & VIB                   & HGP                   & MO                    & RGC                   & MVG                   & GCA                   \\ \hline

        \multirow{8}{*}{\textbf{SN}}        & \multirow{2}{*}{\textbf{I-B}}           & \textit{Acc}                        & 68.40                 & 71.00                 & 68.60                 & 72.80                 & 72.40                 & 71.50                 & \underline{74.30}     & 68.60                 & \textbf{74.40}        & 69.80                 & 71.50                 & 72.90                 & 70.70                 & 72.80                 & 72.50                 & 70.30                 & 73.10                 & 62.30                 & 69.70                 & 71.20                 \\
                                            &                                         & \textit{F1}                         & 69.63                 & 71.17                 & 69.91                 & 73.91                 & 73.25                 & 71.40                 & 72.98                 & 71.05                 & \textbf{76.25}        & 71.46                 & 71.58                 & 73.91                 & 70.14                 & 72.93                 & 71.85                 & 70.78                 & \underline{74.22}     & 63.21                 & 69.40                 & 71.47                 \\ \cline{2-23}

                                            & \multirow{2}{*}{\textbf{I-M}}           & \textit{Acc}                        & 46.20                 & 48.20                 & 46.50                 & 49.80                 & 49.53                 & 49.73                 & \textbf{51.67}        & 46.70                 & \underline{50.80}     & 47.50                 & 47.27                 & 49.67                 & OOM                   & 49.33                 & 47.20                 & 46.50                 & 50.67                 & 41.30                 & 48.67                 & 48.33                 \\
                                            &                                         & \textit{F1}                         & 43.93                 & 47.06                 & 45.12                 & 48.42                 & 47.96                 & 48.22                 & \textbf{49.86}        & 44.35                 & 48.50                 & 45.63                 & 44.66                 & 48.39                 & OOM                   & 48.02                 & 44.43                 & 43.76                 & \underline{48.74}     & 38.67                 & 47.88                 & 47.26                 \\ \cline{2-23}

                                            & \multirow{2}{*}{\textbf{RED}}           & \textit{Acc}                        & \textbf{93.05}        & 89.65                 & 90.94                 & OOM                   & OOM                   & OOM                   & OOM                   & \underline{92.80}     & 91.95                 & 92.60                 & OOM                   & OOM                   & OOM                   & 85.85                 & 82.76                 & OOM                   & 86.25                 & OOM                   & OOM                   & OOM                   \\
                                            &                                         & \textit{F1}                         & \textbf{93.22}        & 90.26                 & 91.32                 & OOM                   & OOM                   & OOM                   & OOM                   & 92.84                 & 92.63                 & \underline{93.03}     & OOM                   & OOM                   & OOM                   & 86.19                 & 82.98                 & OOM                   & 87.22                 & OOM                   & OOM                   & OOM                   \\ \cline{2-23}

                                            & \multirow{2}{*}{\textbf{COL}}           & \textit{Acc}                        & 76.44                 & 73.28                 & 74.06                 & OOM                   & 80.54                 & 79.72                 & \textbf{82.90}        & 75.56                 & 81.64                 & 76.96                 & OOM                   & OOM                   & OOM                   & 80.20                 & 75.28                 & 69.88                 & \underline{82.78}     & OOM                   & 76.88                 & OOM                   \\
                                            &                                         & \textit{F1}                         & 74.54                 & 70.99                 & 71.69                 & OOM                   & 77.79                 & 76.56                 & \underline{80.00}     & 73.95                 & 79.73                 & 75.20                 & OOM                   & OOM                   & OOM                   & 77.45                 & 71.14                 & 67.02                 & \textbf{80.46}        & OOM                   & 73.20                 & OOM                   \\ \hline

        \multirow{11}{*}{\textbf{BIO}}      & \multirow{2}{*}{\textbf{PRO}}           & \textit{Acc}                        & 72.86                 & 72.50                 & 73.67                 & 73.72                 & 72.72                 & 71.07                 & 74.13                 & 72.77                 & 74.40                 & 72.86                 & 70.61                 & \textbf{74.75}        & 70.71                 & 73.85                 & 73.22                 & 73.04                 & 72.32                 & 70.15                 & 73.67                 & 73.22                 \\
                                            &                                         & \textit{F1}                         & 65.96                 & 64.32                 & 66.17                 & 66.45                 & 65.69                 & 63.74                 & 64.84                 & 67.12                 & \textbf{70.58}        & 65.98                 & 58.38                 & 69.78                 & 60.26                 & 63.84                 & 66.41                 & 66.53                 & 64.42                 & 63.86                 & \underline{69.93}     & 66.47                 \\ \cline{2-23}

                                            & \multirow{2}{*}{\textbf{DD}}            & \textit{Acc}                        & 73.10                 & 72.33                 & 75.42                 & 74.88                 & OOM                   & OOM                   & OOM                   & 71.56                 & \textbf{78.02}        & 73.43                 & OOM                   & \underline{77.76}     & 73.35                 & 75.22                 & 76.32                 & 75.98                 & 76.32                 & OOM                   & OOM                   & OOM                   \\
                                            &                                         & \textit{F1}                         & 66.76                 & 65.55                 & 67.07                 & 72.45                 & OOM                   & OOM                   & OOM                   & 63.84                 & \underline{72.52}     & 66.87                 & OOM                   & \textbf{72.67}        & 64.02                 & 70.42                 & 68.45                 & 68.41                 & 68.44                 & OOM                   & OOM                   & OOM                   \\ \cline{2-23}

                                            & \multirow{2}{*}{\textbf{ENZ}}           & \textit{Acc}                        & 46.50                 & 48.33                 & 51.83                 & 52.33                 & 47.83                 & 46.67                 & \textbf{55.50}        & 48.00                 & 49.50                 & 47.83                 & 46.17                 & 52.67                 & 46.50                 & 46.67                 & 44.17                 & 46.67                 & 52.50                 & 49.17                 & \underline{53.17}     & 48.17                 \\
                                            &                                         & \textit{F1}                         & 47.75                 & 47.81                 & 50.98                 & 52.04                 & 47.92                 & 46.80                 & \textbf{54.65}        & 47.94                 & 42.37                 & 47.37                 & 45.92                 & 52.79                 & 46.28                 & 46.22                 & 43.37                 & 46.95                 & 53.09                 & 49.02                 & \underline{53.14}     & 47.72                 \\ \cline{2-23}

                                            & \multirow{2}{*}{\textbf{HIV}}           & \textit{Acc}                        & 96.96                 & 97.01                 & 96.95                 & 97.03                 & 97.01                 & 96.97                 & 97.03                 & 96.87                 & 96.91                 & 96.96                 & 96.89                 & \textbf{97.28}        & 97.03                 & 96.96                 & 96.86                 & 96.89                 & 97.01                 & 96.86                 & \underline{97.05}     & 97.01                 \\
                                            &                                         & \textit{F1}                         & 31.92                 & 34.66                 & 29.91                 & 36.22                 & 32.53                 & 32.38                 & 35.20                 & 22.21                 & 28.61                 & 32.13                 & 25.37                 & \textbf{36.82}        & 34.13                 & 28.04                 & 22.03                 & 25.18                 & 25.02                 & 22.25                 & \underline{36.28}     & 34.48                 \\ \cline{2-23}

                                            & \multirow{3}{*}{\textbf{TOX}}           & \textit{Acc}                        & 55.48                 & 55.53                 & 55.29                 & 55.61                 & 55.49                 & 55.30                 & 55.57                 & 55.23                 & 54.12                 & 55.14                 & 54.63                 & 55.61                 & \textbf{55.80}        & 55.54                 & 52.83                 & 53.50                 & 53.84                 & 51.12                 & \underline{55.65}     & 55.58                 \\
                                            &                                         & \textit{Mi-F1}                      & 91.20                 & 91.14                 & 90.96                 & 91.35                 & 91.17                 & 90.90                 & \textbf{91.40}        & 90.83                 & 91.09                 & 91.01                 & 91.21                 & \underline{91.38}     & 91.37                 & 91.35                 & 89.93                 & 90.04                 & 90.48                 & 89.86                 & 91.26                 & 91.23                 \\
                                            &                                         & \textit{Ma-F1}                      & 36.28                 & 36.01                 & 34.18                 & 38.39                 & 36.79                 & 33.12                 & \textbf{38.61}        & 22.40                 & 34.52                 & 37.65                 & \underline{38.58}     & 38.50                 & 38.15                 & 37.10                 & 20.20                 & 21.25                 & 21.80                 & 19.81                 & 38.47                 & 36.95                 \\ \hline

        \multirow{9}{*}{\textbf{CHE}}       & \multirow{2}{*}{\textbf{MUT}}           & \textit{Acc}                        & 80.41                 & 85.70                 & 81.99                 & 85.88                 & 86.26                 & 82.60                 & \textbf{89.39}        & 80.94                 & 82.54                 & 80.91                 & 80.91                 & 84.09                 & 80.62                 & 82.02                 & 76.43                 & 77.06                 & 77.90                 & 70.67                 & 86.07                 & \underline{88.42}     \\
                                            &                                         & \textit{F1}                         & 85.83                 & 89.07                 & 86.44                 & 89.09                 & 89.24                 & 86.75                 & \textbf{92.38}        & 85.90                 & 86.42                 & 85.06                 & 85.58                 & 87.70                 & 85.12                 & 86.40                 & 82.55                 & 82.96                 & 84.47                 & 80.12                 & 89.12                 & \underline{90.70}     \\ \cline{2-23}

                                            & \multirow{2}{*}{\textbf{NCI}}           & \textit{Acc}                        & 81.58                 & 81.54                 & 81.46                 & 81.83                 & 81.80                 & 80.12                 & \textbf{82.92}        & 81.69                 & 76.62                 & 81.60                 & 78.66                 & \underline{81.87}     & 76.03                 & 81.56                 & 78.16                 & 78.25                 & 78.52                 & 69.95                 & 75.60                 & 81.22                 \\
                                            &                                         & \textit{F1}                         & 81.60                 & 81.53                 & 81.49                 & 81.84                 & 81.28                 & 80.19                 & \textbf{82.85}        & 81.71                 & 77.08                 & 81.68                 & 79.14                 & \underline{81.90}     & 76.73                 & 81.75                 & 78.27                 & 78.31                 & 78.55                 & 69.97                 & 75.57                 & 81.19                 \\ \cline{2-23}

                                            & \multirow{2}{*}{\textbf{BAC}}           & \textit{Acc}                        & 67.11                 & 66.45                 & 64.77                 & 70.74                 & 68.64                 & 67.32                 & \textbf{71.71}        & 67.56                 & 65.13                 & 67.49                 & \underline{71.25}     & 68.03                 & 60.78                 & 67.54                 & 60.19                 & 66.89                 & 62.94                 & 60.89                 & 70.99                 & 67.29                 \\
                                            &                                         & \textit{F1}                         & 71.54                 & 73.51                 & 70.99                 & 73.42                 & 73.72                 & 71.80                 & \underline{74.26}     & 72.64                 & 73.66                 & \textbf{74.36}        & 74.00                 & 73.82                 & 70.55                 & 72.32                 & 70.01                 & 65.45                 & 70.77                 & 70.47                 & 73.66                 & 71.86                 \\ \cline{2-23}

                                            & \multirow{3}{*}{\textbf{PCB}}           & \textit{Acc}                        & 54.56                 & 54.66                 & 54.51                 & 54.70                 & 54.20                 & 54.50                 & \underline{55.01}     & 54.61                 & 54.47                 & TLE                   & OOM                   & \textbf{55.23}        & 54.14                 & 54.77                 & OOM                   & OOM                   & OOM                   & OOM                   & OOM                   & OOM                   \\
                                            &                                         & \textit{Mi-F1}                      & 98.50                 & 98.53                 & 98.39                 & 98.53                 & 98.50                 & 96.45                 & \underline{98.56}     & 98.49                 & 98.36                 & TLE                   & OOM                   & \textbf{98.60}        & 98.50                 & 98.53                 & OOM                   & OOM                   & OOM                   & OOM                   & OOM                   & OOM                   \\
                                            &                                         & \textit{Ma-F1}                      & 12.00                 & 15.00                 & 13.17                 & 15.17                 & 11.83                 & 12.83                 & 18.20                 & 13.33                 &  1.67                 & TLE                   & OOM                   & \textbf{21.40}        & \underline{20.64}     & 17.33                 & OOM                   & OOM                   & OOM                   & OOM                   & OOM                   & OOM                   \\ \hline
    \end{tabular}
    \label{tab:graph_classification}
    \vspace{-0.5em}
\end{table*}

\begin{table*}
    \caption{
		Evaluation on graph regression. 
		The best and second-best results are highlighted in bold and underlined, respectively.
	}
    \centering
    \setlength\tabcolsep{0.96pt}
    \renewcommand{\arraystretch}{1.1}
    \begin{tabular}{|c|c|c|c|c|c|c|c|c|c|c|c|c|c|c|c|c|c|c|c|c|c|c|}
        \hline
        \multirow{2}{*}{\textbf{Type}}      & \multirow{2}{*}{\textbf{Data}}          & \multirow{2}{*}{\textit{\textbf{Met.}}} & \multicolumn{7}{c|}{\textbf{Node-based}}                                                                                                                                                                           & \multicolumn{3}{c|}{\textbf{Pooling-based}}                                  & \multicolumn{4}{c|}{\textbf{Subgraph-based}}                                                   & \multicolumn{3}{c|}{\textbf{GL-based}}                                           & \multicolumn{3}{c|}{\textbf{SSL-based}}                                          \\ \cline{4-23} 
        
                                            &                                         &                                         & GCN                   & GIN                   & SAGE                  & PNA                   & NAG                   & HGT                   & GPS                   & TopK                  & GMT                   & EP                 & ECS                   & AK+                   & I2                    & HMN                  & VIB                   & HGP                   & MO                    & RGC                   & MVG                   & GCA                   \\ \hline

        \multirow{28}{*}{\textbf{MC}}       & \multirow{2}{*}{\textbf{3-Cyc}}         & \textit{MAE}                            & 0.440                 & 0.396                 & 0.512                 & 0.406                 & 0.360                 & 0.375                 & 0.023                 & 0.429                 & 0.425                 & 0.424                 & 0.019                 & \underline{0.002}     & \textbf{0.001}        & 0.036                 & 0.878                 & 0.437                 & 0.541                 & 0.474                 & 0.423                 & 0.387                 \\
                                            &                                         & \textit{R2}                             & 0.697                 & 0.755                 & 0.808                 & 0.749                 & 0.794                 & 0.777                 & 0.998                 & 0.704                 & 0.711                 & 0.568                 & 1.000                 & \underline{1.000}     & \textbf{1.000}        & 0.994                 & 0.103                 & 0.685                 & 0.500                 & 0.606                 & 0.717                 & 0.760                 \\ \cline{2-23}

                                            & \multirow{2}{*}{\textbf{4-Cyc}}         & \textit{MAE}                            & 0.281                 & 0.254                 & 0.541                 & 0.251                 & 0.275                 & 0.278                 & 0.034                 & 0.277                 & 0.272                 & 0.270                 & \underline{0.015}     & 0.022                 & \textbf{0.006}        & 0.041                 & 0.645                 & 0.275                 & 0.544                 & 0.540                 & 0.273                 & 0.220                 \\
                                            &                                         & \textit{R2}                             & 0.823                 & 0.886                 & 0.401                 & 0.892                 & 0.837                 & 0.840                 & 0.997                 & 0.833                 & 0.848                 & 0.841                 & \underline{1.000}     & 0.999                 & \textbf{1.000}        & 0.996                 & 0.142                 & 0.835                 & 0.403                 & 0.444                 & 0.861                 & 0.899                 \\ \cline{2-23}

                                            & \multirow{2}{*}{\textbf{5-Cyc}}         & \textit{MAE}                            & 0.278                 & 0.186                 & 0.461                 & 0.258                 & 0.205                 & 0.266                 & 0.069                 & 0.240                 & 0.234                 & 0.266                 & 0.072                 & \underline{0.034}     & \textbf{0.012}        & 0.117                 & 0.902                 & 0.276                 & 0.453                 & 0.457                 & 0.218                 & 0.176                 \\
                                            &                                         & \textit{R2}                             & 0.833                 & 0.936                 & 0.482                 & 0.887                 & 0.881                 & 0.857                 & 0.988                 & 0.861                 & 0.894                 & 0.855                 & 0.986                 & \underline{0.997}     & \textbf{1.000}        & 0.968                 & 0.002                 & 0.819                 & 0.608                 & 0.494                 & 0.906                 & 0.941                 \\ \cline{2-23}

                                            & \multirow{2}{*}{\textbf{6-Cyc}}         & \textit{MAE}                            & 0.301                 & 0.190                 & 0.469                 & 0.284                 & 0.187                 & 0.265                 & 0.064                 & 0.230                 & 0.179                 & 0.293                 & 0.086                 & \underline{0.058}     & \textbf{0.037}        & 0.120                 & 0.888                 & 0.304                 & 0.456                 & 0.503                 & 0.177                 & 0.178                 \\
                                            &                                         & \textit{R2}                             & 0.793                 & 0.916                 & 0.619                 & 0.825                 & 0.920                 & 0.831                 & 0.985                 & 0.880                 & 0.928                 & 0.807                 & 0.957                 & \underline{0.996}     & \textbf{0.997}        & 0.942                 & 0.002                 & 0.789                 & 0.622                 & 0.608                 & 0.930                 & 0.922                 \\ \cline{2-23}

                                            & \multirow{2}{*}{\textbf{7-Cyc}}         & \textit{MAE}                            & 0.401                 & 0.211                 & 0.590                 & 0.224                 & 0.220                 & 0.324                 & 0.059                 & 0.285                 & 0.157                 & 0.394                 & 0.156                 & \underline{0.056}     & \textbf{0.049}        & 0.114                 & 0.827                 & 0.422                 & 0.571                 & 0.584                 & 0.150                 & 0.205                 \\
                                            &                                         & \textit{R2}                             & 0.589                 & 0.864                 & 0.459                 & 0.847                 & 0.853                 & 0.676                 & 0.990                 & 0.792                 & 0.935                 & 0.609                 & 0.946                 & \underline{0.994}     & \textbf{0.995}        & 0.968                 & 0.006                 & 0.549                 & 0.488                 & 0.469                 & 0.940                 & 0.877                 \\ \cline{2-23}

                                            & \multirow{2}{*}{\textbf{8-Cyc}}         & \textit{MAE}                            & 0.476                 & 0.263                 & 0.529                 & 0.284                 & 0.292                 & 0.433                 & 0.053                 & 0.291                 & 0.138                 & 0.470                 & 0.115                 & \underline{0.049}     & \textbf{0.040}        & 0.149                 & 0.743                 & 0.479                 & 0.534                 & 0.481                 & 0.129                 & 0.259                 \\
                                            &                                         & \textit{R2}                             & 0.357                 & 0.722                 & 0.221                 & 0.715                 & 0.750                 & 0.629                 & 0.994                 & 0.755                 & 0.943                 & 0.384                 & 0.970                 & \underline{0.995}     & \textbf{0.996}        & 0.952                 & 0.033                 & 0.352                 & 0.205                 & 0.332                 & 0.947                 & 0.727                 \\ \cline{2-23}

                                            & \multirow{2}{*}{\textbf{4-Path}}        & \textit{MAE}                            & 0.715                 & 0.427                 & 0.734                 & 0.294                 & 0.398                 & 0.448                 & 0.017                 & 0.527                 & 0.161                 & 0.636                 & 0.024                 & \underline{0.015}     & \textbf{0.008}        & 0.075                 & 0.778                 & 0.502                 & 0.732                 & 0.727                 & 0.151                 & 0.409                 \\
                                            &                                         & \textit{R2}                             & 0.157                 & 0.635                 & 0.090                 & 0.862                 & 0.670                 & 0.615                 & 0.999                 & 0.527                 & 0.946                 & 0.319                 & 0.992                 & \underline{1.000}     & \textbf{1.000}        & 0.990                 & 0.028                 & 0.539                 & 0.091                 & 0.131                 & 0.953                 & 0.658                 \\ \cline{2-23}

                                            & \multirow{2}{*}{\textbf{5-Path}}        & \textit{MAE}                            & 0.685                 & 0.395                 & 0.667                 & 0.363                 & 0.389                 & 0.419                 & 0.016                 & 0.519                 & 0.156                 & 0.613                 & \underline{0.013}     & 0.015                 & \textbf{0.009}        & 0.080                 & 0.751                 & 0.485                 & 0.693                 & 0.705                 & 0.141                 & 0.381                 \\
                                            &                                         & \textit{R2}                             & 0.105                 & 0.636                 & 0.168                 & 0.697                 & 0.647                 & 0.607                 & 0.999                 & 0.502                 & 0.947                 & 0.325                 & \underline{1.000}     & \underline{1.000}     & \textbf{1.000}        & 0.982                 & 0.028                 & 0.520                 & 0.082                 & 0.065                 & 0.956                 & 0.656                 \\ \cline{2-23}

                                            & \multirow{2}{*}{\textbf{6-Path}}        & \textit{MAE}                            & 0.616                 & 0.391                 & 0.659                 & 0.340                 & 0.361                 & 0.423                 & \underline{0.013}     & 0.563                 & 0.139                 & 0.693                 & 0.014                 & \underline{0.013}     & \textbf{0.009}        & 0.070                 & 0.757                 & 0.455                 & 0.663                 & 0.659                 & 0.130                 & 0.378                 \\
                                            &                                         & \textit{R2}                             & 0.247                 & 0.596                 & 0.129                 & 0.743                 & 0.636                 & 0.597                 & 0.999                 & 0.413                 & 0.954                 & 0.141                 & 1.000                 & \underline{1.000}     & \textbf{1.000}        & 0.989                 & 0.025                 & 0.546                 & 0.128                 & 0.128                 & 0.961                 & 0.608                 \\ \cline{2-23}

                                            & \multirow{2}{*}{\textbf{4-Cliq}}        & \textit{MAE}                            & 0.343                 & 0.345                 & 0.350                 & 0.250                 & 0.343                 & 0.345                 & 0.014                 & 0.343                 & 0.343                 & 0.358                 & \underline{0.009}     & 0.009                 & \textbf{0.001}        & 0.010                 & 0.387                 & 0.236                 & 0.342                 & 0.342                 & 0.180                 & 0.340                 \\
                                            &                                         & \textit{R2}                             & 0.161                 & 0.135                 & 0.109                 & 0.823                 & 0.160                 & 0.137                 & 0.932                 & 0.121                 & 0.102                 & 0.106                 & 0.967                 & \underline{0.996}     & \textbf{1.000}        & \underline{0.996}     & 0.089                 & 0.890                 & 0.160                 & 0.161                 & 0.900                 & 0.164                 \\ \cline{2-23}

                                            & \multirow{2}{*}{\textbf{\begin{tabular}[c]{@{}c@{}}Tailed\\ Tri\end{tabular}}} & \textit{MAE}  & 0.351    & 0.289                 & 0.381                 & 0.347                 & 0.258                 & 0.299                 & 0.019                 & 0.347                 & 0.367                 & 0.341                 & 0.019                 & \underline{0.015}     & \textbf{0.003}        & 0.064                 & 0.893                 & 0.379                 & 0.389                 & 0.410                 & 0.328                 & 0.275                 \\
                                            &                                         & \textit{R2}                             & 0.788                 & 0.861                 & 0.730                 & 0.779                 & 0.888                 & 0.846                 & 0.999                 & 0.781                 & 0.767                 & 0.805                 & 0.999                 & \underline{1.000}     & \textbf{1.000}        & 0.989                 & 0.018                 & 0.734                 & 0.701                 & 0.625                 & 0.818                 & 0.866                 \\ \cline{2-23}

                                            & \multirow{2}{*}{\textbf{\begin{tabular}[c]{@{}c@{}}Chor.\\ Cyc\end{tabular}}}  & \textit{MAE}  & 0.438    & 0.368                 & 0.439                 & 0.350                 & 0.347                 & 0.368                 & 0.036                 & 0.422                 & 0.369                 & 0.425                 & \underline{0.030}     & 0.034                 & \textbf{0.004}        & 0.104                 & 0.872                 & 0.444                 & 0.441                 & 0.473                 & 0.367                 & 0.361                 \\
                                            &                                         & \textit{R2}                             & 0.635                 & 0.711                 & 0.600                 & 0.812                 & 0.738                 & 0.714                 & 0.994                 & 0.641                 & 0.685                 & 0.642                 & \underline{0.997}     & 0.995                 & \textbf{1.000}        & 0.921                 & 0.033                 & 0.575                 & 0.584                 & 0.568                 & 0.729                 & 0.715                 \\ \cline{2-23}

                                            & \multirow{2}{*}{\textbf{\begin{tabular}[c]{@{}c@{}}Tri.\\ Rec.\end{tabular}}}  & \textit{MAE}  & 0.466    & 0.403                 & 0.479                 & 0.441                 & 0.396                 & 0.427                 & 0.348                 & 0.474                 & 0.407                 & 0.489                 & 0.362                 & \underline{0.346}     & \textbf{0.327}        & 0.400                 & 0.826                 & 0.429                 & 0.476                 & 0.541                 & 0.430                 & 0.394                 \\
                                            &                                         & \textit{R2}                             & 0.521                 & 0.629                 & 0.534                 & 0.560                 & 0.620                 & 0.593                 & \underline{0.708}     & 0.492                 & 0.624                 & 0.470                 & 0.690                 & 0.706                 & \textbf{0.724}        & 0.632                 & 0.034                 & 0.583                 & 0.504                 & 0.385                 & 0.569                 & 0.643                 \\ \hline
    \end{tabular}
    \label{tab:graph_regression}
\end{table*}

\subsubsection{Graph Classification Tasks}\label{ssec:exp:classification}
As shown in  Table~\ref{tab:graph_classification},
node-based GNNs, such as GCN and SAGE, cannot achieve satisfactory performance. 
These models learn node representations and use global pooling.
Global pooling overlooks local structural information, which is critical for distinguishing graphs in bioinformatics (e.g., ENZYMES) and chemistry (e.g., MUTAG). 
PNA achieves comparatively good performance as it captures richer neighbor distributions by leveraging multiple aggregators and degree-scalers. 
Also, 
Graph Transformers (GPS, NAG, HGT) capture long-range dependencies effectively 
but lack explicit motif extraction, limiting their performance on datasets like 
PROTEINS that rely on fine-grained substructures.
Secondly, 
pooling-based approaches, such as GMT and TopK, achieve competitive performance, particularly on social networks like COLLAB and REDDIT. 
These methods progressively coarsen the graph by grouping or selecting the most 
important nodes, preserving multi-level structural information.
However, they are less effective on datasets where fine-grained local structures (e.g., motifs) play a critical role, such as ENZYMES and NCI1. 

Subgraph-based methods,  such as ECS, AK+, and I2,  demonstrate highly competitive performance on bioinformatics and chemistry datasets because they break graphs into meaningful substructures, 
enabling them to capture important patterns that other methods often miss. 
Within this category, HMN improves efficiency by sampling only a 
few subgraphs via walk centrality, but this aggressive reduction loses local 
structural detail, limiting its generalization compared to exhaustive methods.
However, they are out-of-memory (OOM) on large graphs or high node counts, such as REDDIT and COLLAB. 
Fourthly, graph learning-based approaches, such as MO and HGP, perform well on noisy social datasets, such as IMDB-B and IMDB-M.
By dynamically reconstructing graph structures and removing irrelevant edges or nodes, these approaches enhance robustness and improve generalization.
However, they are less effective on datasets where the original graph structure is already well-formed molecular graphs, such as MUTAG.
Lastly, SSL-based methods like MVG and GCA achieve robust performance across 
multiple datasets by pretraining on unlabeled graphs, though their graph 
augmentation overhead can lead to OOM issues.

\subsubsection{Graph Regression Tasks}\label{ssec:exp:regression}
For graph regression tasks, the focus is primarily on evaluating how well GNNs can capture the semantics and key structural patterns of graphs, such as cycle and path counts for each graph. 
Lower \textit{MAE} and higher \textit{$R^2$} scores reflect better performance. 
As shown in Table~\ref{tab:graph_regression}, node-based methods (excluding GIN and PNA), pooling-based models, 
GL-based techniques, and SSL-based approaches generally fail to deliver satisfactory results. 
This is because these methods are not specifically designed to enhance the expressiveness of GNNs, meaning they cannot effectively differentiate between isomorphic graphs or graphs with identical cycles. 
In contrast, GIN, PNA, and subgraph-based GNNs explicitly aim to improve the theoretical expressiveness of GNNs, resulting in superior performance. 
GIN and PNA aim to improve the theoretical expressiveness to approximate the Weisfeiler-Lehman (1-WL) isomorphism test. 
Similarly, although GTs (e.g., GPS, NAG, HGT) show improved capabilities over standard node-based methods in capturing broader contexts, they still lack the strict theoretical expressivity guarantees required to precisely count complex structural motifs.

Subgraph-based models like ECS, AK+, and I2 consistently outperform other approaches on almost all regression datasets. By breaking graphs down into overlapping or rooted subgraphs, these methods capture rich structural details that enhance their theoretical expressivity. As a result, they are better able to distinguish between graph isomorphism classes and accurately identify important motifs, which leads to improved performance. 
Notably, because HMN makes deliberate compromises for computational speed, its capacity to distinguish the topological roles of multi-hop neighbors is diminished. Consequently, its performance on counting tasks that emphasize complex exact structures falls slightly behind other subgraph-based models.
Lastly, 
as the complexity of the target motif increases (e.g., from 3-Cycle to 8-Cycle), most GNNs (except subgraph-based ones) experience a drastic performance drop. 
This highlights the limitations of many GNN architectures in capturing higher-order dependencies and complex structural relationships.

\subsection{Efficiency Evaluation}\label{ssec:efficiency}

\begin{figure*}[t]
	\centering

	{\centering\includegraphics[width=0.8\linewidth]{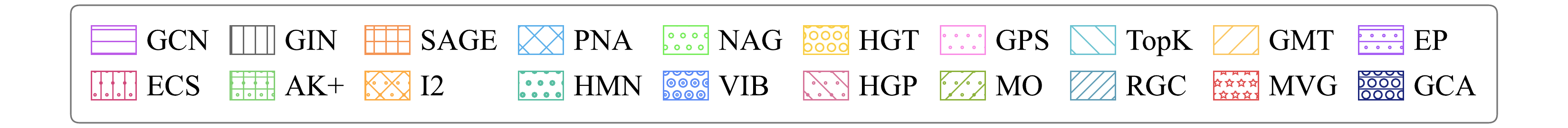}}

	\vspace{-0.5em}

	
	\subfloat[{\vspace{-1em}ENZYMES dataset.}]
	{\centering\includegraphics[width=1\linewidth]{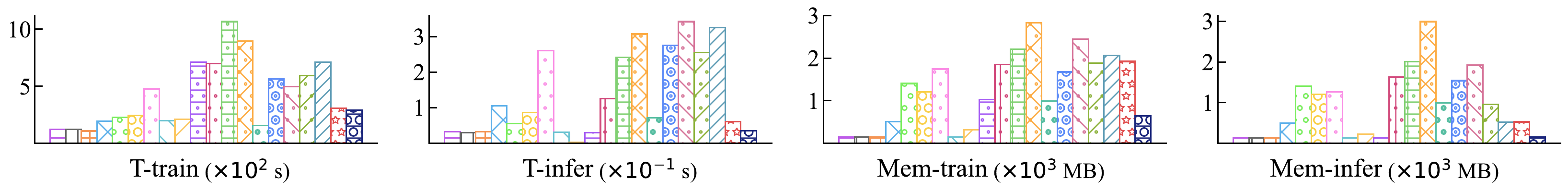}}

	\vspace{-0.5em}

	\subfloat[{MUTAG dataset.}]
	{\centering\includegraphics[width=1\linewidth]{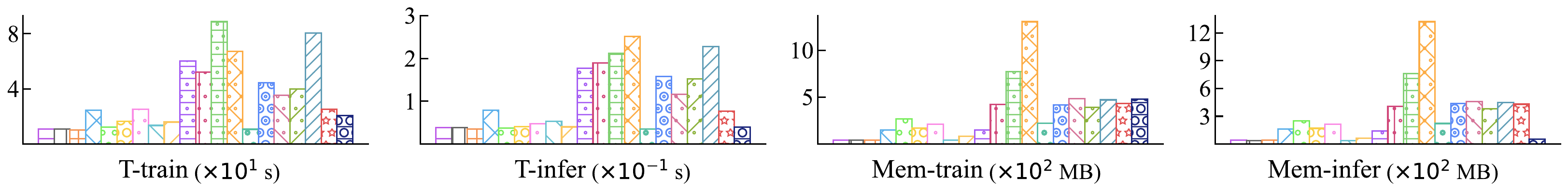}}


	\caption{Efficiency evaluation.}
	
	\label{fig:efficiency}
	\vspace{-1em}
\end{figure*}
	\begin{figure*}[t]
	\centering
	\vspace{-1em}
	\subfloat[Accuracy]{
		\includegraphics[width=0.1905\linewidth]{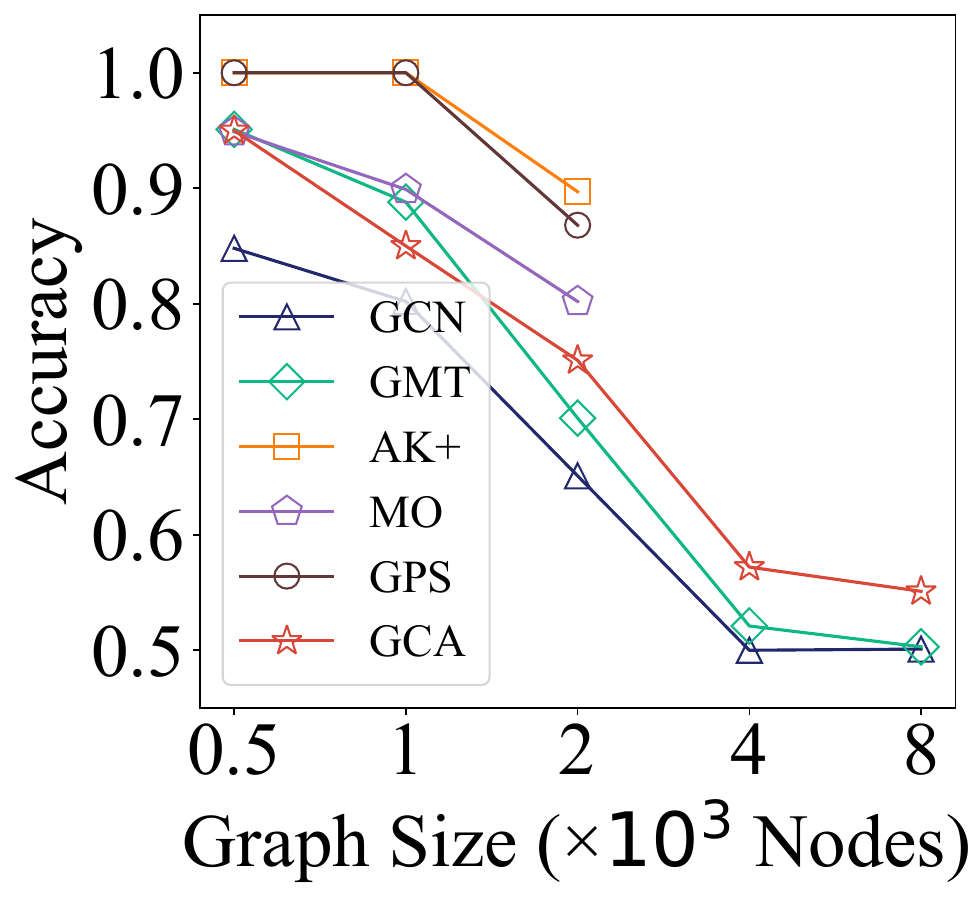}
		\label{fig:scalability_acc}
	}%
	\hfill
	\subfloat[Train Memory]{
		\includegraphics[width=0.187\linewidth]{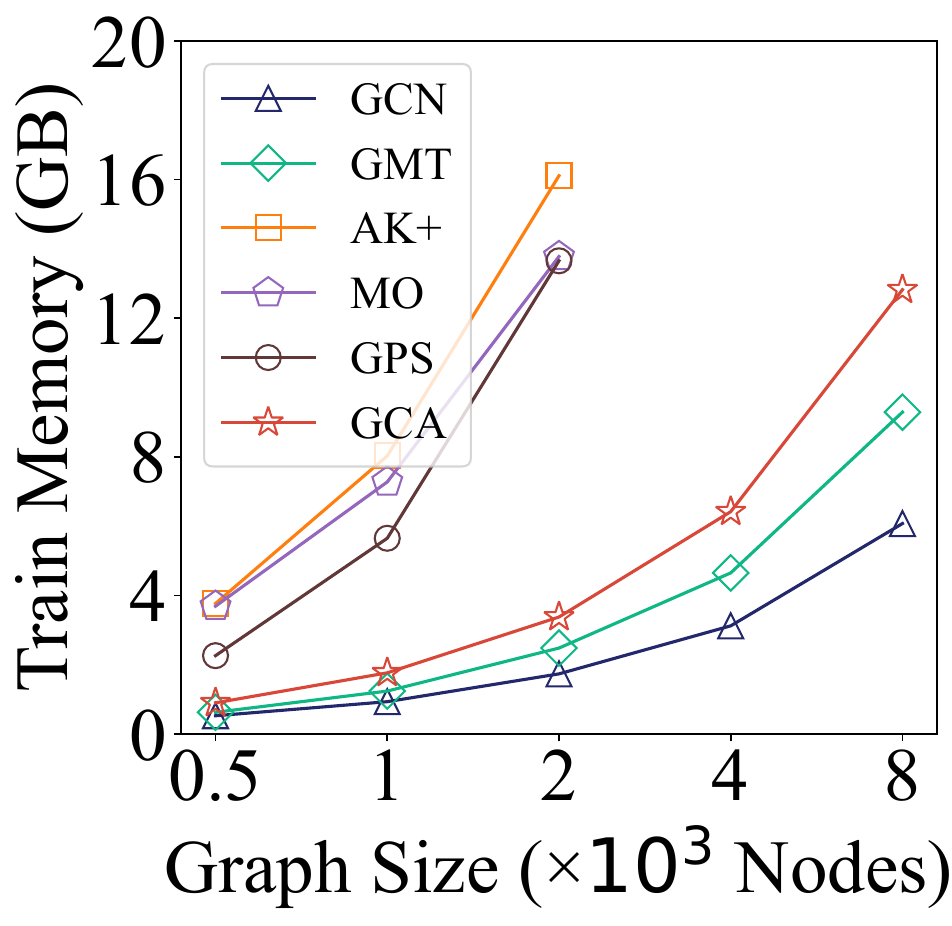}
		\label{fig:scalability_train_mem}
	}%
	\hfill
	\subfloat[Train Time]{
		\includegraphics[width=0.183\linewidth]{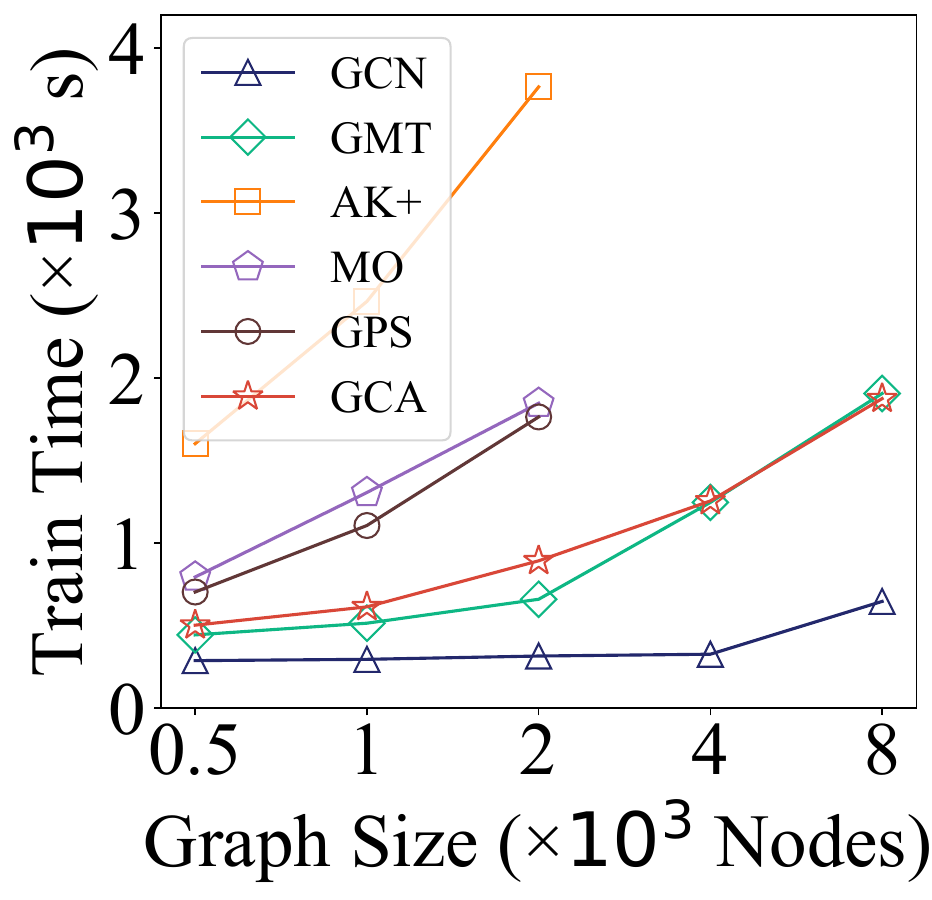}
		\label{fig:scalability_train_time}
	}%
	\hfill
	\subfloat[Inference Memory]{
		\includegraphics[width=0.187\linewidth]{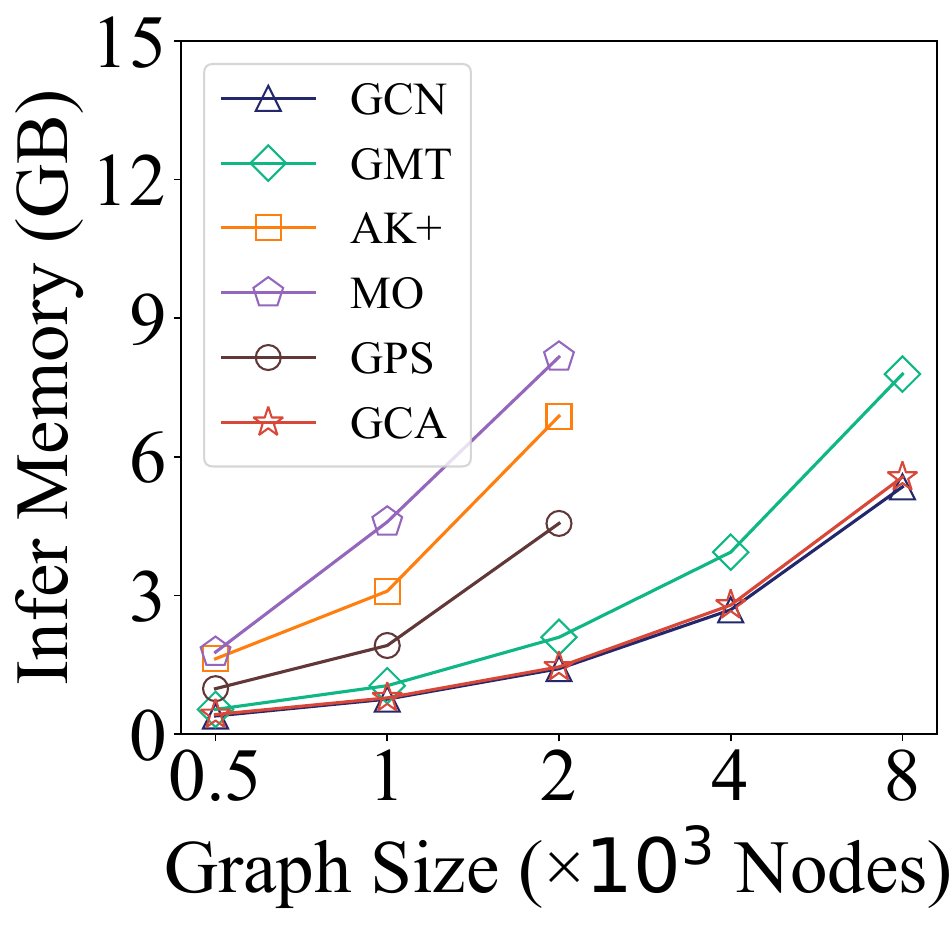}
		\label{fig:scalability_infer_mem}
	}%
	\hfill
	\subfloat[Inference Time]{
		\includegraphics[width=0.18\linewidth]{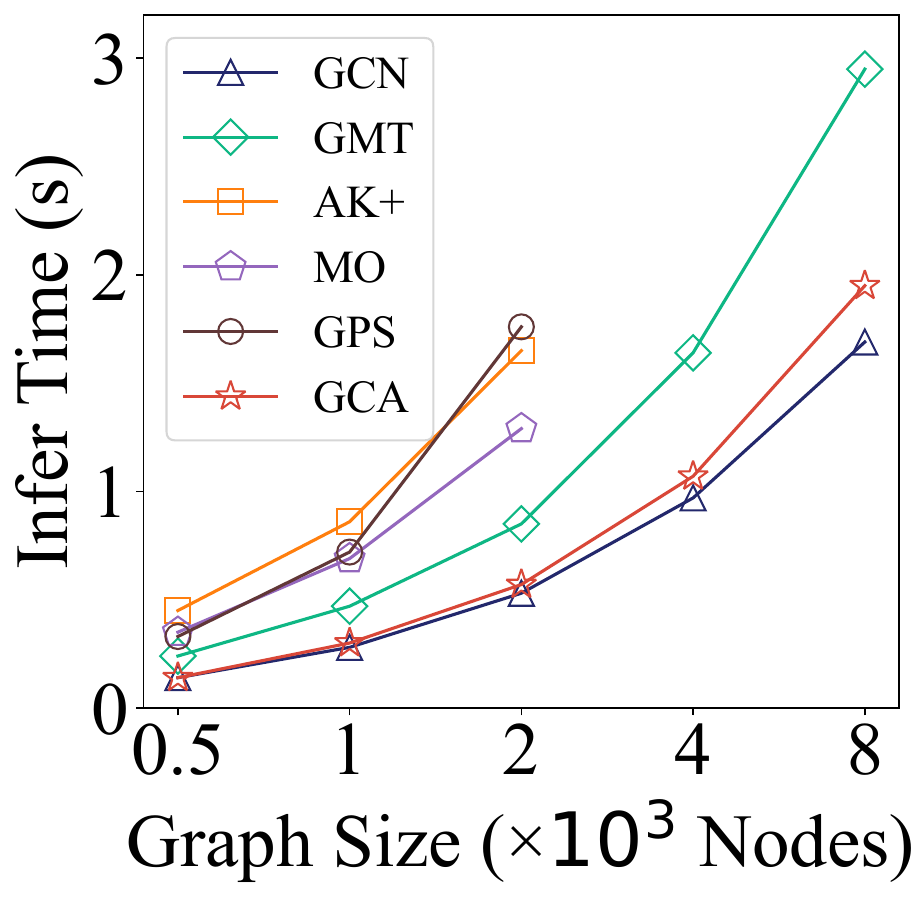}
		\label{fig:scalability_infer_time}
	}
	
	\caption{Scalability evaluation with different graph sizes.}
	\label{fig:scalability}
	\vspace{-1em}
\end{figure*}

We evaluate the efficiency of GNNs in terms of total training time, inference time, GPU peak memory usage during training and inference. 
As shown in Figure~\ref{fig:efficiency}, node-based GNNs exhibit the highest efficiency across all datasets. 
Their simplicity in aggregating neighbors’ information without additional graph processing ensures minimal time and memory usage. 
However, this efficiency comes at the cost of limited expressiveness as discussed in Section~\ref{ssec:effectiveness}. 
Among node-based variants, GTs introduce substantial computational overhead. GPS incurs severe memory usage and scalability bottlenecks due to its quadratic global attention. 
NAG and HGT effectively mitigate the high resource demands of standard GTs, benefiting from NAG’s mini-batch training via Hop2Token and HGT’s completely decoupled graph processing via hub labeling.
Secondly, 
pooling-based methods strike a balance between efficiency and performance. TopK is efficient due to its node-pruning strategy, which reduces graph size while preserving key features. However, more advanced pooling approaches, such as GMT, are computationally expensive due to their clustering operations based on node similarity. 
Thirdly, 
subgraph-based GNNs (e.g., ECS, AK+, I2) are computationally intensive because they rely on generating multiple subgraphs for each graph. Despite their high resource requirements, these methods excel at capturing graph motifs and complex structural patterns, as shown in Section~\ref{ssec:effectiveness}. Their computational cost makes them less efficient for real-time or large-scale applications.
An exception within this category is HMN, which circumvents the expensive memory and computational overhead of exhaustive subgraph methods with walk-based centrality sampling.
GL-based methods (e.g., VIB, HGP, and MO) dynamically reconstruct graph structures during training, which adds significant computational overhead. While they improve robustness to noise and graph quality, their iterative optimization process limits scalability to larger datasets.
SSL-based methods (e.g., RGC, MVG, GCA) are computationally expensive during training because they require graph augmentations and contrastive learning stage.

In summary, node-based GNNs are the most efficient but lack expressiveness, while pooling-based models strike a balance between efficiency and performance. Subgraph-based and GL-based approaches offer superior expressiveness but suffer from high computational costs. SSL-based methods, though computationally expensive during training, are efficient during inference, making them suitable for pretraining scenarios.

\subsection{Scalability Evaluations}\label{ssec:scalability}
To evaluate the memory and computational scalability of GNNs on larger graphs, 
we construct a stress-test environment using a single RTX 3090 GPU (24\,GB). 
We adopt the synthetic BA2Motifs benchmark~\cite{luo2020parameterized}, a 
binary classification task that distinguishes sparse Barab\'{a}si--Albert (BA) 
graphs by the presence of a ``House'' motif. Because the generation rule 
maintains a nearly constant average degree across all scales, resource 
consumption in this experiment reflects graph size rather than structural 
density~\cite{ying2019gnnexplainer}.

Whereas conventional graph-level datasets such as IMDB and MolTox21 
predominantly contain very small graphs—typically 20 to 100 nodes as shown 
in Table~\ref{tab:data}—our setup probes substantially larger scales. We fix 
the batch size at 16 and systematically increase the node count across 
$\{500,\,1000,\,2000,\,4000,\,8000\}$. To ensure a comprehensive yet 
representative comparison, we select six top-performing models, one from each 
architectural category in Table~\ref{tab:graph_classification}: GCN, GPS, GMT, 
AK+, MO, and GCA. Under configurations identical to our primary experiments, 
we run 10-fold cross-validation at each scale and record average test accuracy 
together with peak allocated GPU memory.

\subsubsection{Classification accuracy.}
Figure~\ref{fig:scalability}(a) shows that all models suffer accuracy drops as 
graph size grows, since the fixed-size motif is increasingly diluted by the 
expanding background during global pooling. GMT and GCA partially alleviate 
this issue—GMT filters out irrelevant nodes through attention-based 
hierarchical pooling, while GCA uses contrastive objectives over augmented 
views to preserve core structural signals. AK+ and GPS maintain accuracy more 
effectively: AK+ captures the motif through localized $k$-hop subgraph 
aggregation, and GPS leverages global attention to relay critical signals 
across the entire graph. MO performs similarly by dynamically pruning 
task-irrelevant structures. However, the higher accuracy of these expressive 
models (AK+, GPS, and MO) comes at the cost of substantial memory and 
computation, limiting their practicality on large graphs.

\subsubsection{Computational resources.}
Figures~\ref{fig:scalability}(b)--(e) highlight stark contrasts in memory and 
time consumption. GCN remains highly efficient, incurring only marginal 
overhead increases as graphs scale. GMT offers a more scalable alternative to 
other expressive architectures, although its hierarchical edge contraction 
risks discarding essential motif structures in larger graphs. In contrast, GPS 
exhibits the most aggressive memory growth, quickly reaching hardware limits 
due to the quadratic complexity of its global attention. AK+ and MO 
also impose substantial costs due to exhaustive subgraph enumeration and 
dynamic graph reconstruction, respectively. GCA has noticeable 
augmentation overhead during training but scales more gracefully than 
dense-attention models.

\begin{figure*}[t]
	\vspace{-1em}
	\centering 	
	\subfloat[IMDB-M]	
	{\centering\includegraphics[width=0.25\linewidth, height=3.05cm]{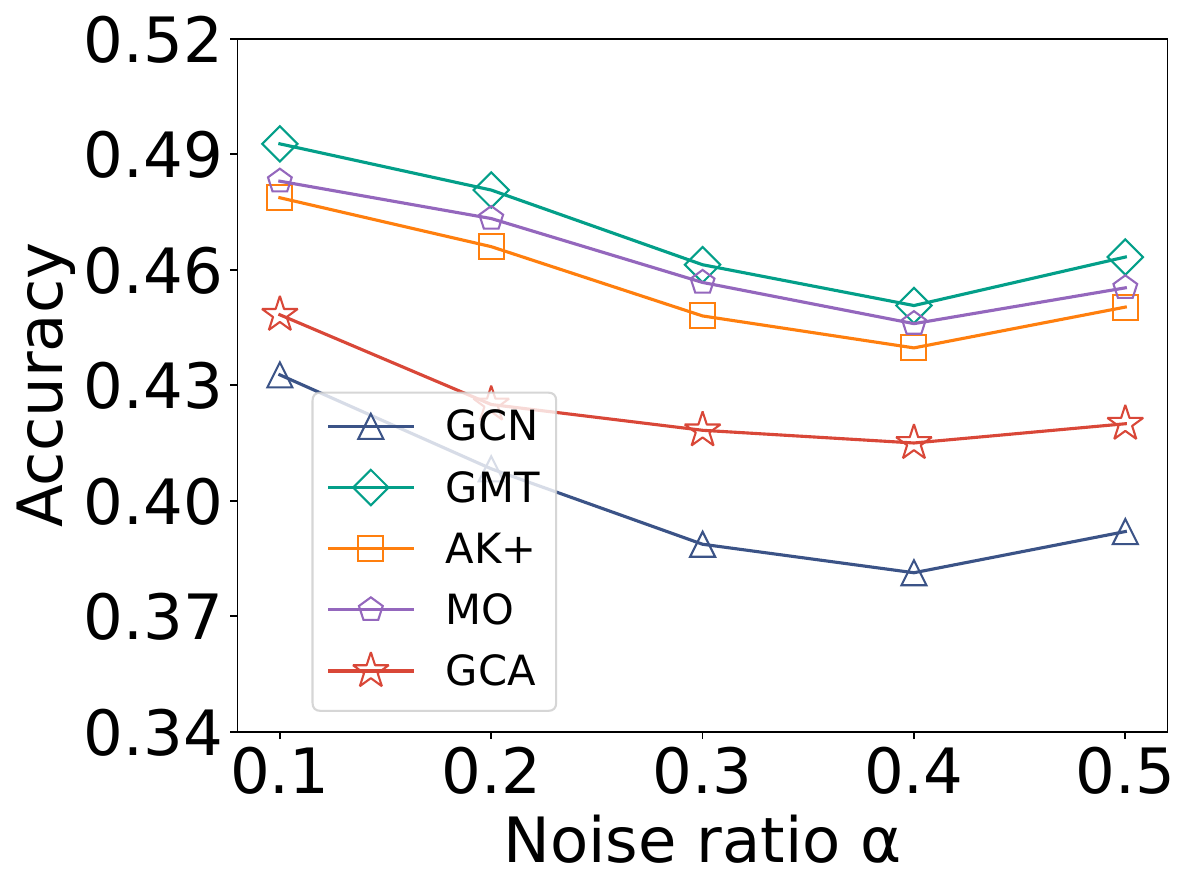}}
	\hfill
	\subfloat[ENZYMES]
	{\centering\includegraphics[width=0.25\linewidth, height=2.97cm]{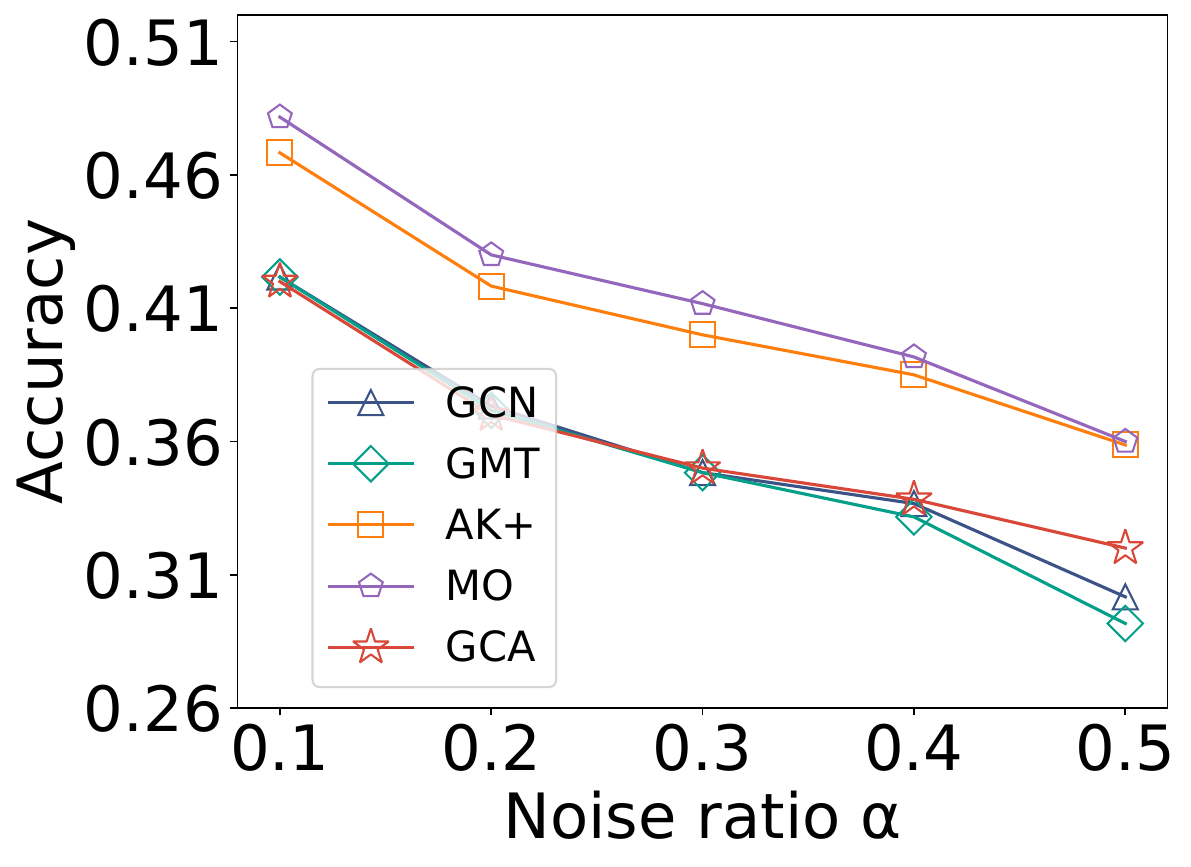}}	
	\subfloat[NCI1]	
	{\centering\includegraphics[width=0.25\linewidth, height=3.05cm]{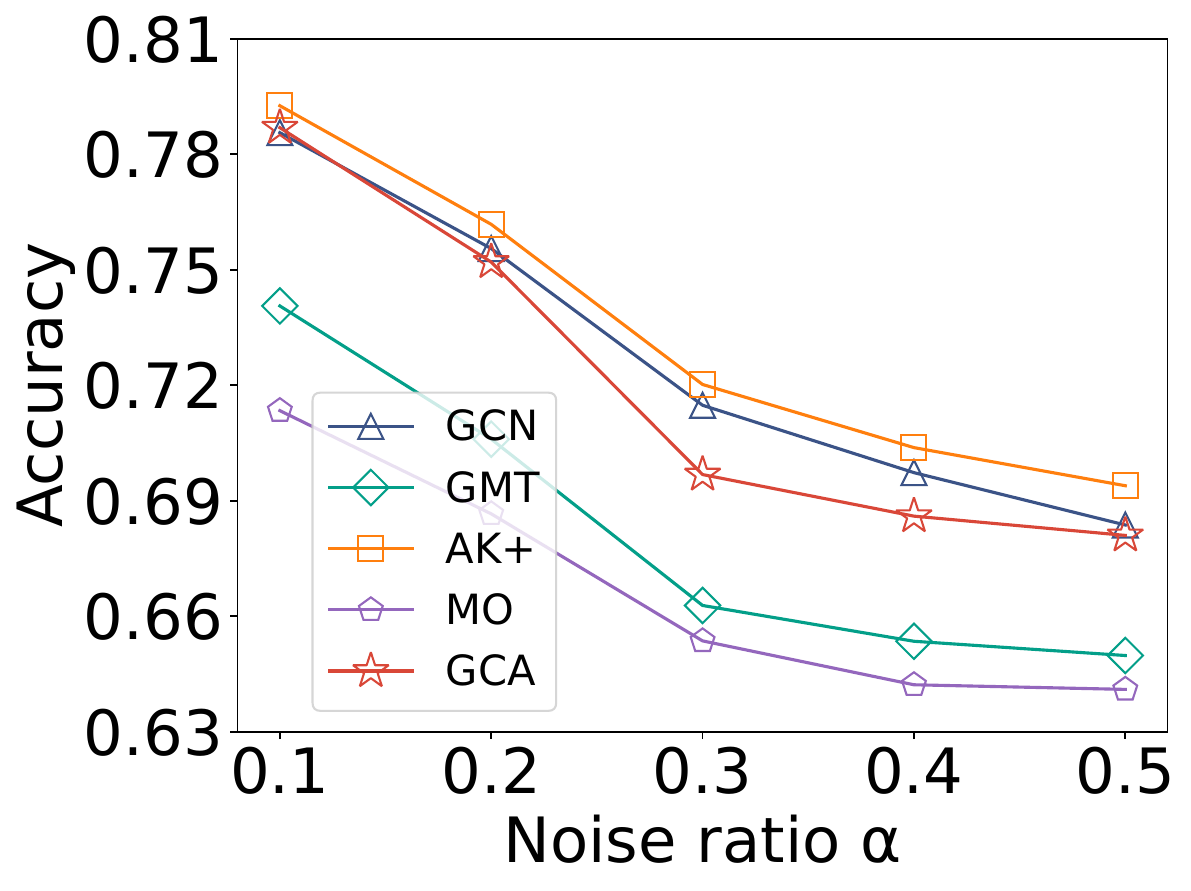}}
	\hfill
	\subfloat[MUTAG]
	{\centering\includegraphics[width=0.25\linewidth, height=3.05cm]{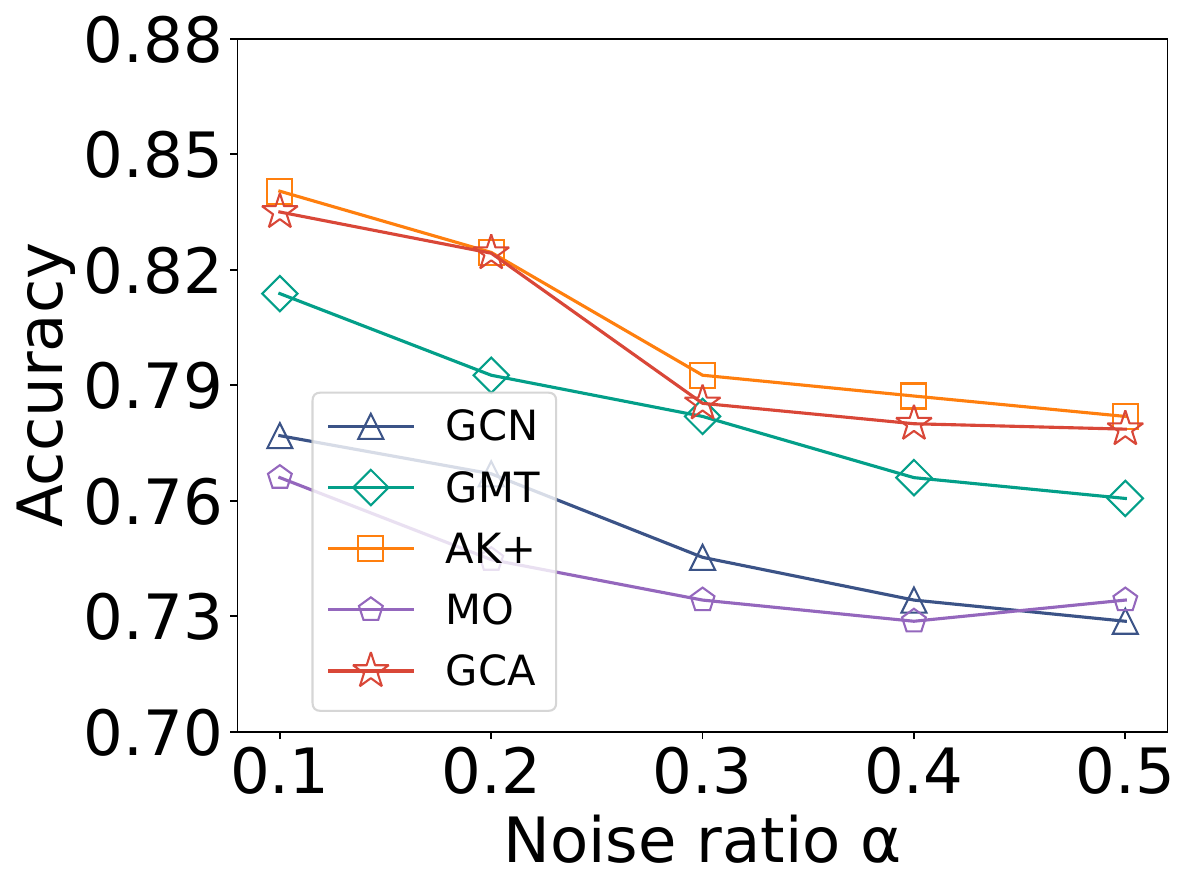}}	
	\hfill
	
	\caption{Robustness evaluation on noisy graphs.}
	\label{fig:robustness}
	\vspace{0em}
\end{figure*}

\begin{figure*}[t]
	\centering 	
	\subfloat[IMDB-M]	
	{\centering\includegraphics[width=0.25\linewidth, height=3.05cm]{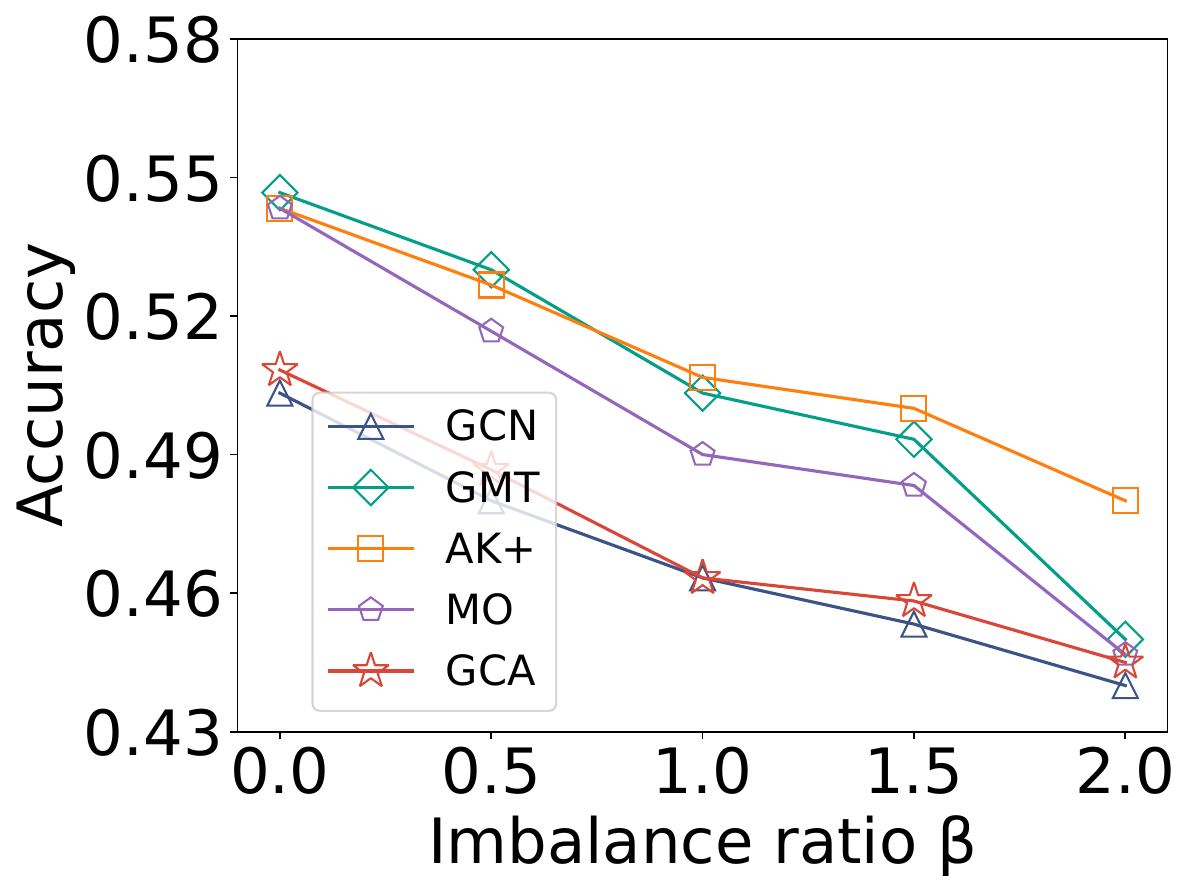}}
	\hfill
	\subfloat[ENZYMES]
	{\centering\includegraphics[width=0.25\linewidth, height=2.97cm]{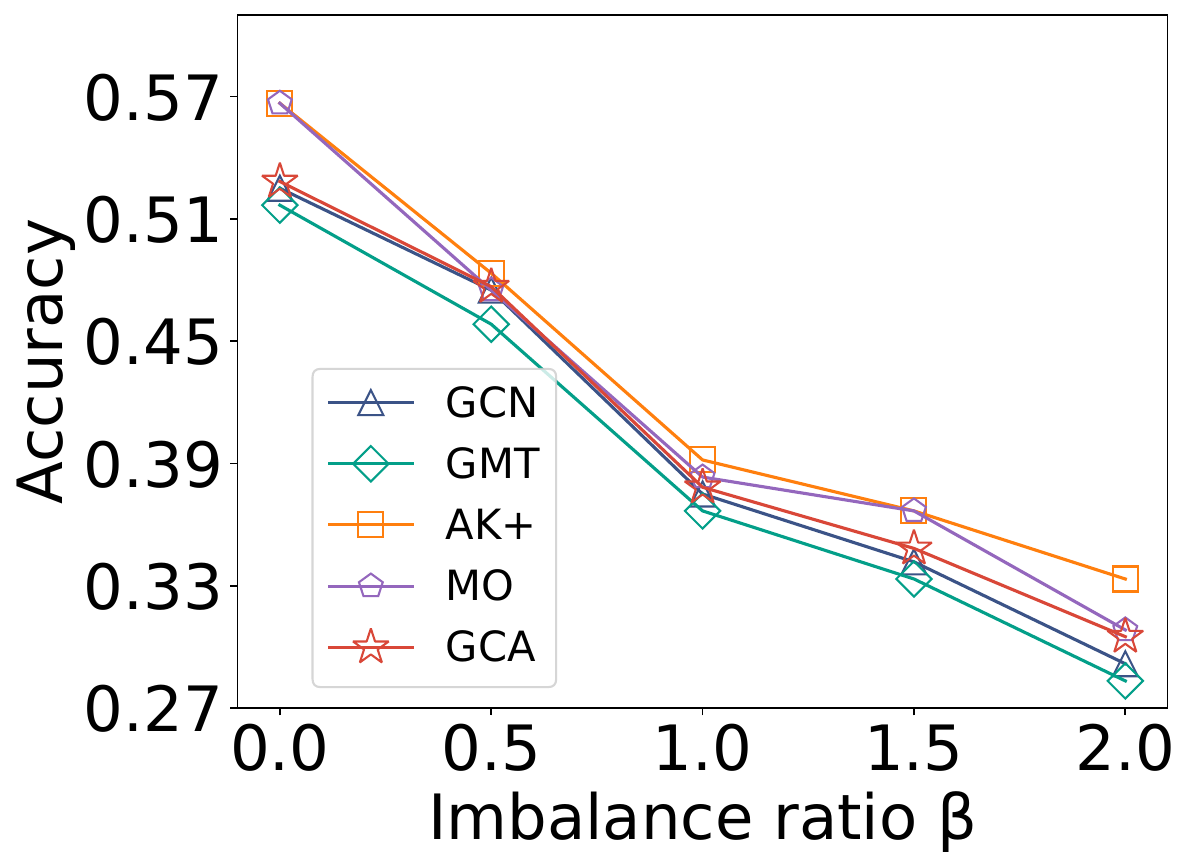}}	
	\subfloat[NCI1]	
	{\centering\includegraphics[width=0.25\linewidth, height=2.98cm]{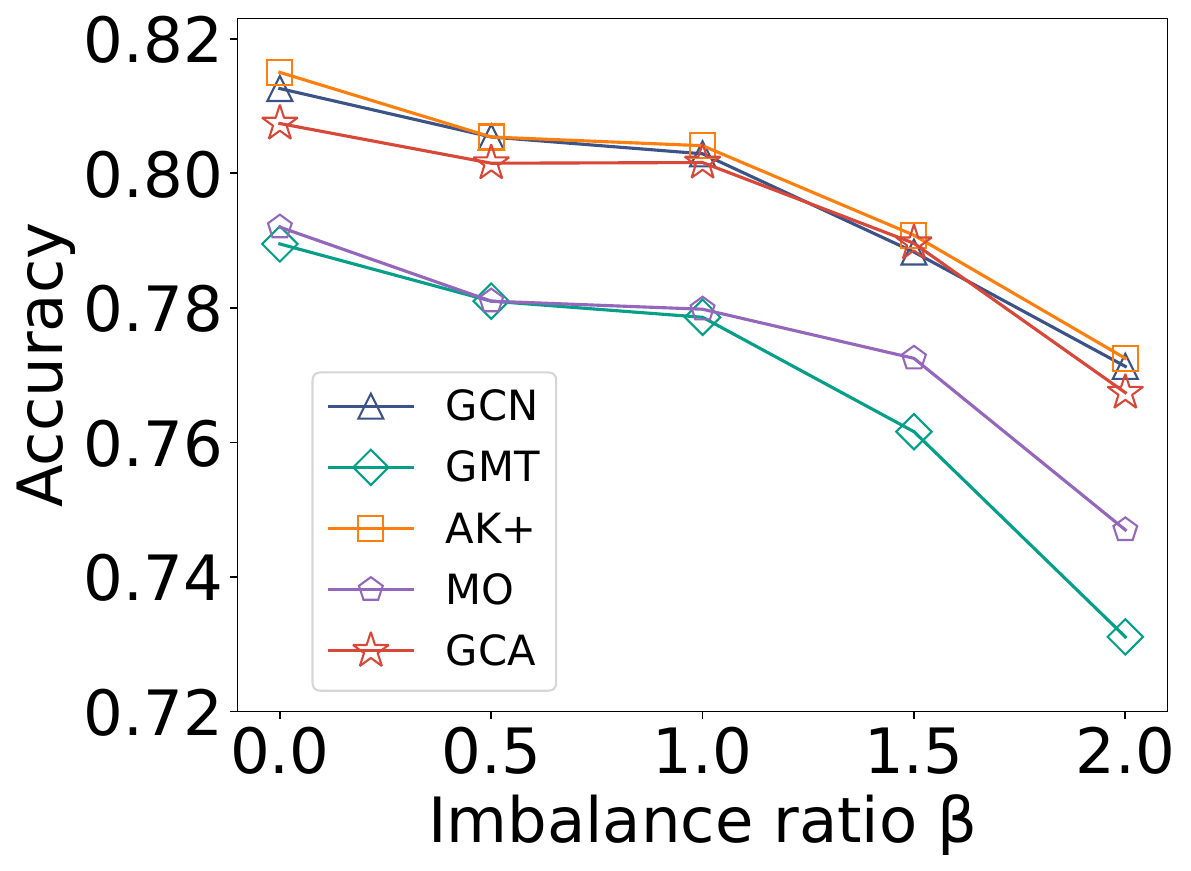}}
	\hfill
	\subfloat[4-Cycle]
	{\centering\includegraphics[width=0.25\linewidth, height=3.05cm]{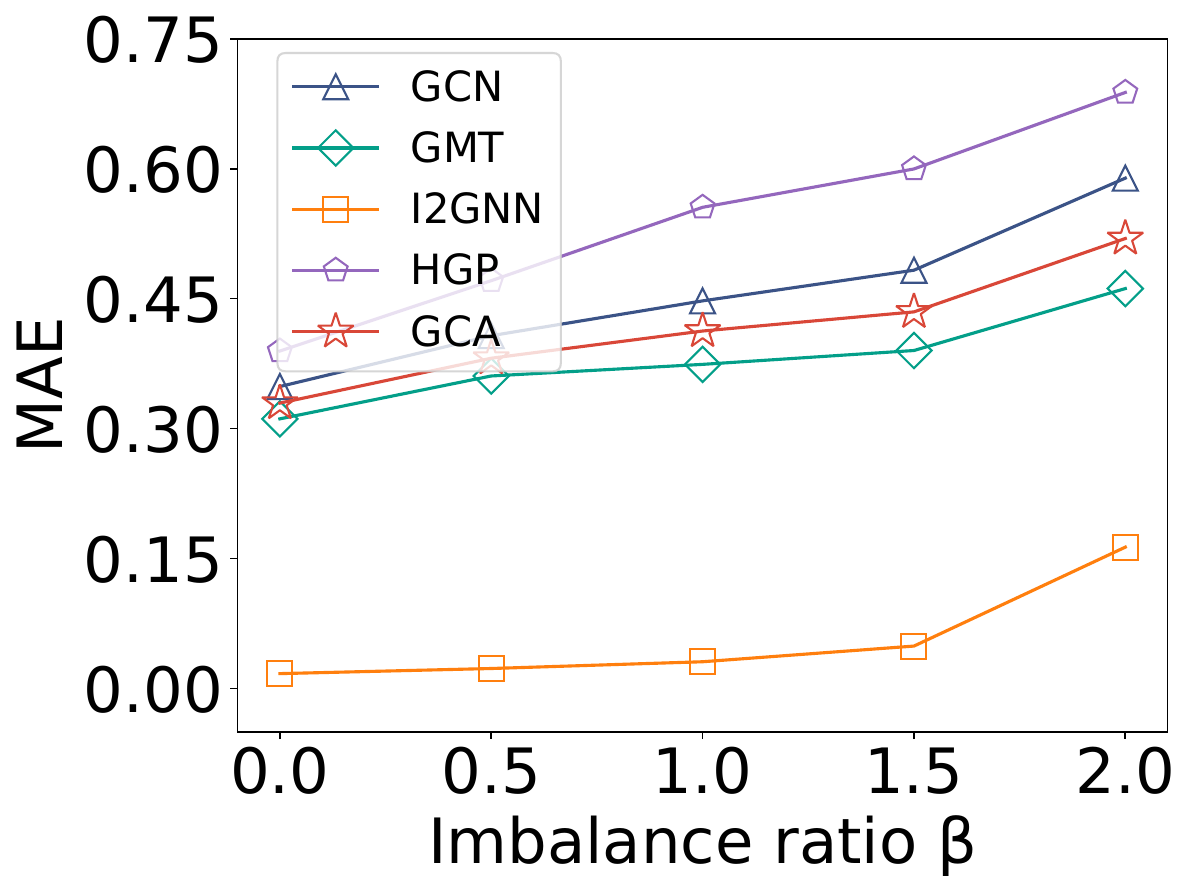}}	
	\hfill
	
	\caption{Imbalance data evaluation.}
	
	\label{fig:imbalance}
	\vspace{0em}
\end{figure*}

\begin{figure*}[t]
	\centering 	
	\subfloat[IMDB-M]	
	{\centering\includegraphics[width=0.249\linewidth, height=3.05cm]{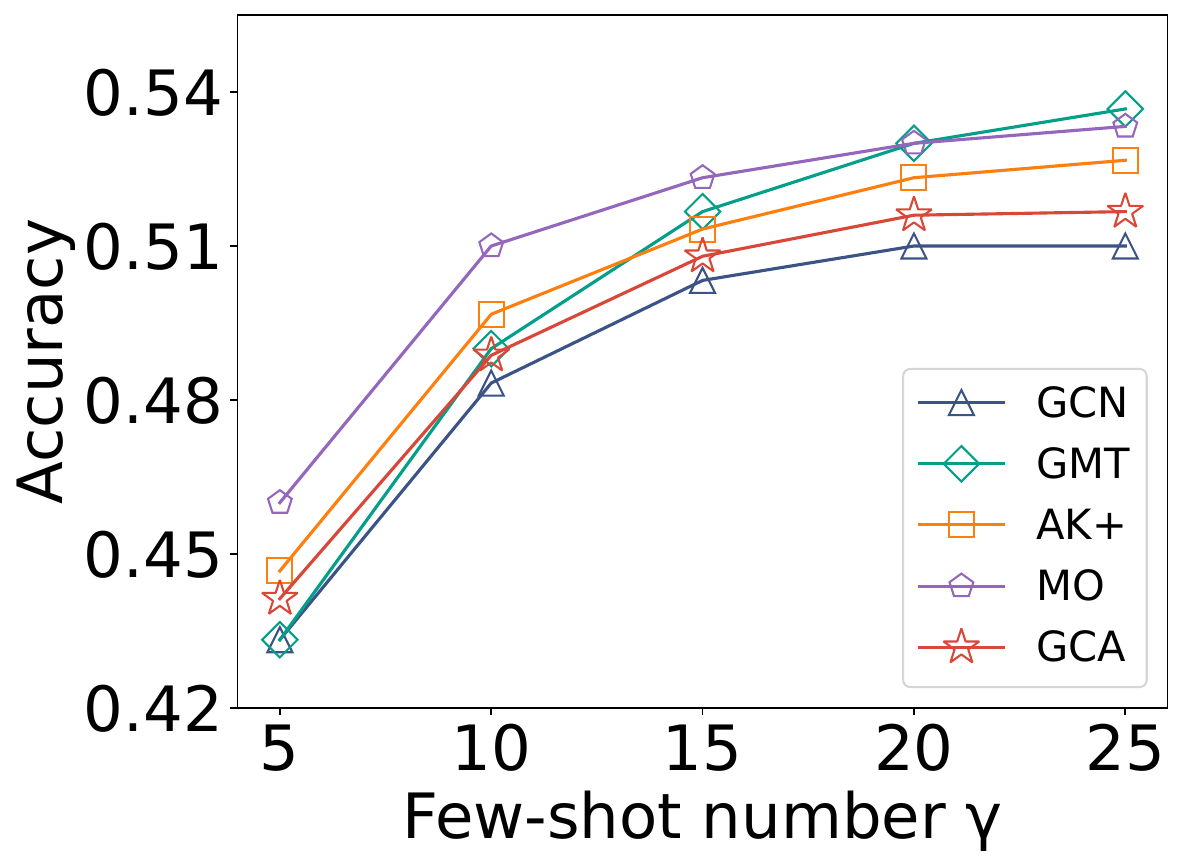}}
	\hfill
	\subfloat[ENZYMES]
	{\centering\includegraphics[width=0.249\linewidth, height=3.05cm]{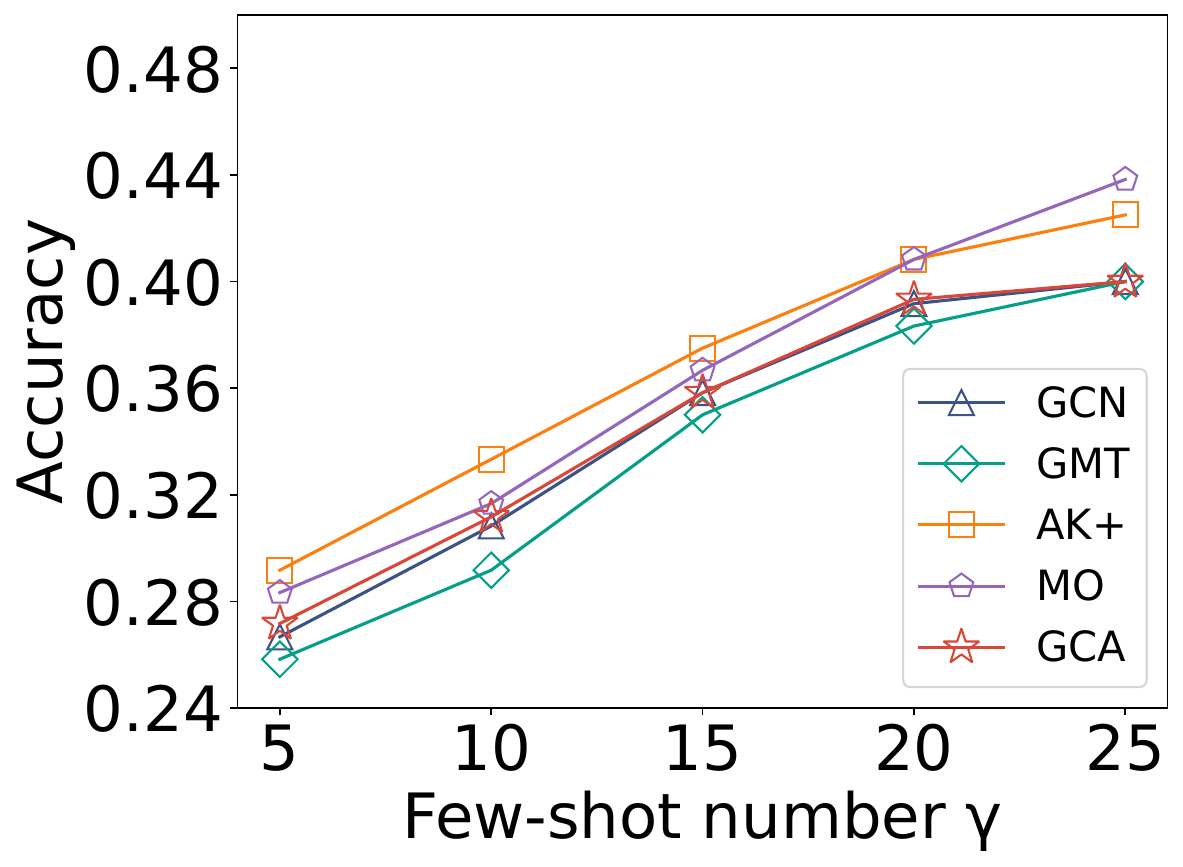}}	
	\subfloat[NCI1]	
	{\centering\includegraphics[width=0.249\linewidth, height=3.12cm]{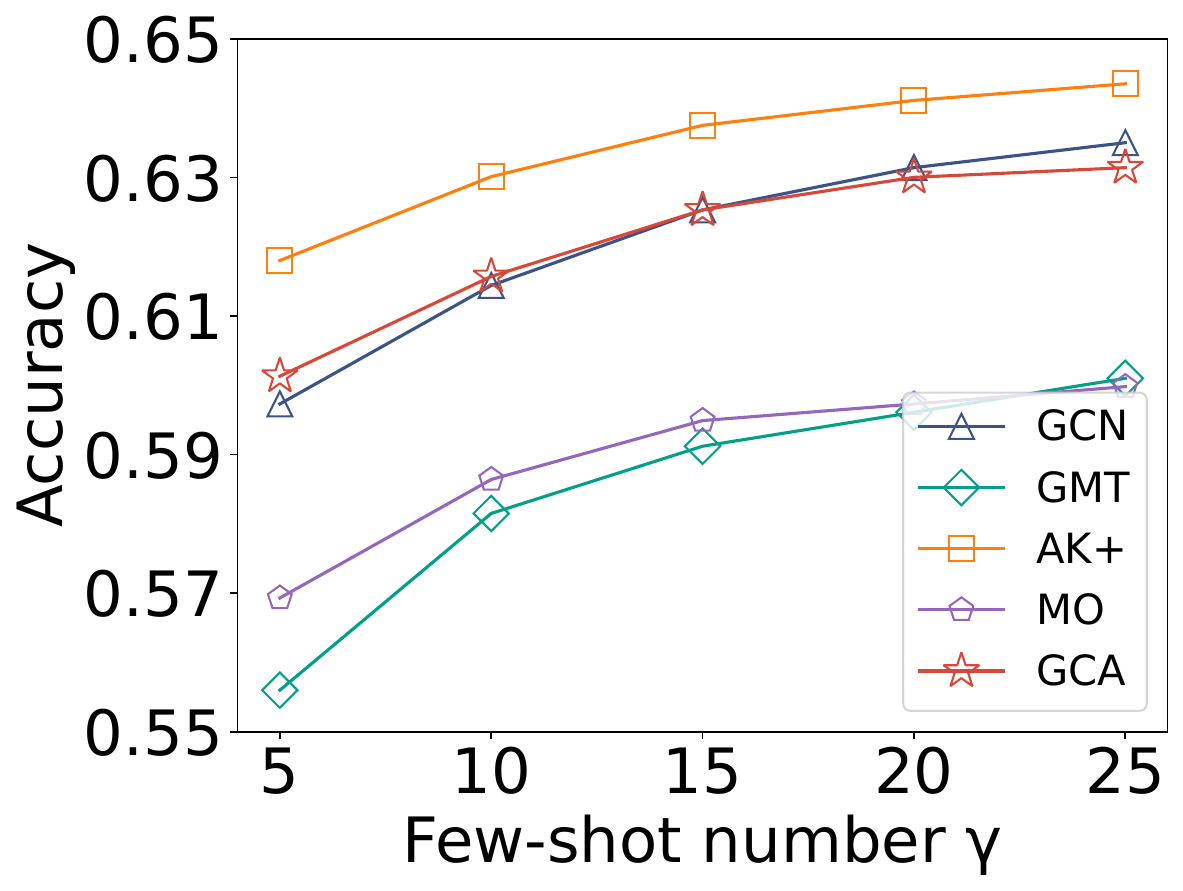}}
	\hfill
	\subfloat[4-Cycle]
	{\centering\includegraphics[width=0.249\linewidth, height=3.12cm]{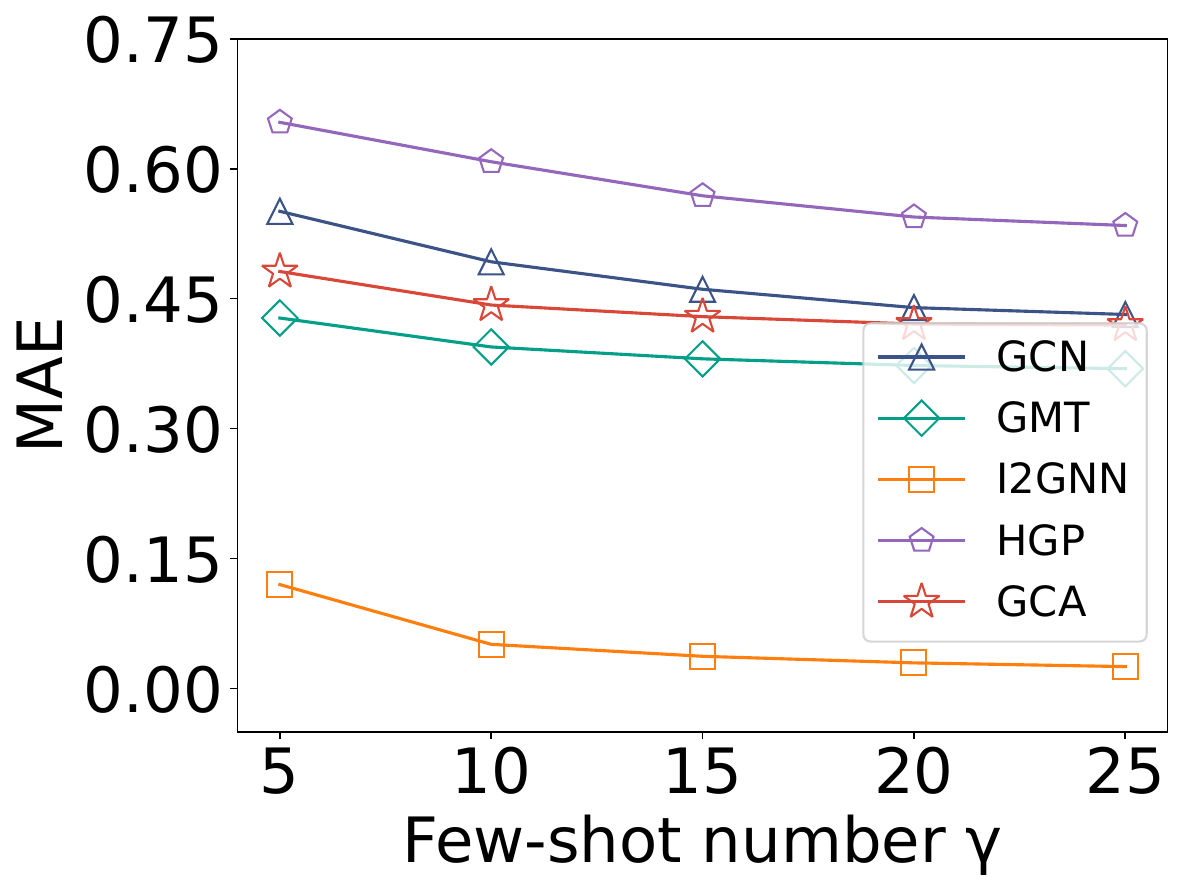}}	
	\hfill
	
	\caption{Few-shot evaluation.}
	\label{fig:few_shot}
\end{figure*}

\subsection{More Scenarios}\label{ssec:scenarios}
We further evaluate GNNs in three realistic and challenging scenarios, including noisy graphs, imbalanced graphs, and few-shot graphs. 
To ensure clarity, we select one representative graph dataset from each domain for evaluation, i.e., IMDB-M, ENZYMES, NCI1, and 4-Cycle. 
Additionally, we compare some of the most representative models from each GNN category, including GCN (node-based), GMT (pooling-based), and GCA (SSL-based). 
For top perform, we select AK+ (subgraph-based), MO (GL-based) for classification tasks, while I2 (subgraph-based) and HGP (GL-based) for regression tasks.

\subsubsection{Robustness Evaluation on Noisy Graphs} 
We assess model performance under structural noise. 
Specifically, for each graph $G(V,\mathbf{A},\mathbf{X})$, we introduce noise by randomly removing $\alpha||\mathbf{A}||_0$ existing edges, where the ratio $\alpha$ takes values in $\{0.1, 0.2, 0.3, 0.4, 0.5\}$.
As shown in Figure~\ref{fig:robustness}, 
as $\alpha$ increases, 
node-based GNNs (e.g., GCN) and Pooling-based models (e.g., GMT) suffer significant performance drops, due to their reliance on the original connectivity for aggregation and clustering. 
In contrast, subgraph-based methods such as AK+ show greater robustness by focusing on small, locally coherent substructures that remain intact despite global noise. 
Graph learning-based models like MO are even more resilient, dynamically reconstructing graph structures to mitigate the impact of noise. 
Similarly, SSL-based methods such as GCA perform well, leveraging augmented views to learn noise-resistant representations. 
Overall, the evidence underscores that subgraph-based, GL-based, and SSL-based methods demonstrate superior robustness to noise compared to node-based and pooling-based models. 
This highlights the practical value of these more sophisticated methods for handling real-world graphs, which are often challenging to model due to inherent noise.

\subsubsection{Imbalance Data Evaluation} 
For the graph classification task, given a dataset $\mathcal{G} = {(G_i, y_i)}$, we simulate class imbalance by adjusting the proportion of training samples in each class to follow the sequence $\{1, \frac{1}{2^\beta}, \frac{1}{3^\beta}, \dots, \frac{1}{|\mathcal{Y}|^\beta}\}$, where $\beta \in \{0, 0.5, 1, 1.5, 2\}$ determines the degree of imbalance. The sample count for the first class remains constant across all $\beta$ settings.
For graph regression tasks, where $y_i$ spans the interval $[y_{\text{min}}, y_{\text{max}}]$, we divide the label range into three equal-sized buckets and set the group proportions as $\{1, \frac{1}{2^\beta}, \frac{1}{3^\beta}\}$ to create imbalance, with larger $\beta$ corresponding to higher level of imbalance.
As shown in Figure~\ref{fig:imbalance}, 
accuracy drops across various GNN architectures as imbalance increases, particularly in datasets with minority classes. 
Node-based GNN (e.g., GCN) and pooling-based GNN (GMT)  struggle because their global aggregation mechanisms tend to average out minority signals, making it difficult to distinguish rare patterns. 
Subgraph-based  (AK+) and graph learning-based (MO) GNNs, while capturing richer structural information, do not explicitly address class imbalance, leading to performance degradation under extreme imbalance, especially for fewer classes. 
SSL-based model (GCA), although capable of leveraging abundant unlabeled data, likewise fails to correct imbalance on its own and converges to performance similar to other baselines unless augmented with imbalance-aware strategies~\cite{liu2023qtiah}.

\subsubsection{Few-shot Evaluation} 
We simulate data scarcity by limiting the number of training samples. 
For classification tasks, we construct training sets where the number of labeled graphs per class is in $\gamma \in \{5, 10, 15, 20, 25\}$. 
For regression tasks, we partition the label into five equal-width buckets and uniformly sample $\gamma$ training instances per bucket. 
The results, as summarized in Figure~\ref{fig:few_shot}, reveal that 
most GNN architectures exhibit significant performance degradation as $\gamma$ decreases. 
Node-based (GCN) and pooling-based (GMT) models are particularly affected, as their global aggregation mechanisms rely heavily on abundant labeled data to learn meaningful representations. 
Interestingly, while subgraph-based models such as AK+ and I2, as well as graph learning-based models like MO and HGP, are theoretically capable of leveraging local structural patterns, they do not demonstrate substantial resilience to data scarcity in practice. 
This is primarily because current implementations lack explicit mechanisms to identify and prioritize the most informative subgraphs or adaptively focus on critical features when labeled data is limited. 
Their performance improvements over standard baselines are thus marginal in few-shot scenarios, indicating that richer local modeling alone does not guarantee data efficiency~\cite{wang2023contrastive}.

\subsection{Correlation Analysis}\label{sec:correlation}

To investigate the intrinsic relationship between graph topology and GNN 
performance, we characterize each graph using \textit{11} topological features 
spanning three groups: global structural features, local structural features, 
and node distributions.

\noindent\textbf{(i) Global Structural Features (6 metrics).}
These metrics provide a high-level summary of overall graph connectivity and 
organization for graph datasets.
(1)~\textit{Average Degree (Deg.)} and 
(2)~\textit{Density (Den.)} measure overall graph sparsity.
(3)~\textit{Average Shortest Path Length (SPL)} and 
(4)~\textit{Diameter (Dia.)} capture the extent and compactness of the 
largest connected component.
(5)~\textit{Degree Assortativity (Assr.)} indicates whether high-degree nodes 
tend to connect with other high-degree nodes.
(6)~\textit{Modularity (Mod.)} quantifies the strength of community structure.

\noindent\textbf{(ii) Local Structural Features (2 metrics).}
These metrics focus on small, recurring substructures.
(7)~\textit{Average Clustering Coefficient (CC)} measures the average density 
of subgraphs induced by a node's neighbors.
(8)~\textit{Triangle Counts (Tri.)} captures the prevalence and heterogeneity 
of dense local motifs.

\noindent\textbf{(iii) Node Distributions (3 metrics).}
These features characterize the distribution of node importance.
(9)~\textit{Betweenness Centrality (BC)} indicates the presence of critical 
bridge nodes controlling information flow.
(10)~\textit{PageRank (PR)} reveals hierarchical structures with distinct 
authority nodes.
(11)~\textit{K-Core Number (KC)} reflects global robustness and core-periphery 
structure.

For metrics defined at the node level, including \textit{Tri.}, \textit{BC.}, \textit{PR.}, \textit{KC.}, 
we report both the \textbf{Mean} ($\mu$) and \textbf{Standard Deviation} ($\sigma$).

\vspace{1ex}
\noindent\textit{Evaluation.}
Given GNNs $\mathcal{M}=\{M_i\}_{i=1}^n$ and graph dataset $\mathcal{D} = \{G_j\}_{j=1}^m$, we construct a performance matrix $\mathbf{P}\in\mathbb{R}^{n\times m}$ where $\mathbf{P}[i][j]$ is the result of $M_i$ on $G_j$ (e.g., accuracy or MAE). 
We denote the $k$-th topological feature of $\mathcal{D}$ as $\mathbf{f}_k$, 
and assess the relationship between model performance and the $k$-th feature with Spearman’s correlation $\rho(\mathbf{P}[i], \mathbf{f}_k)$, using a significance threshold of $\alpha = 0.05$.

\begin{figure}[t]
	\centering 	
	\includegraphics[width=1\linewidth]{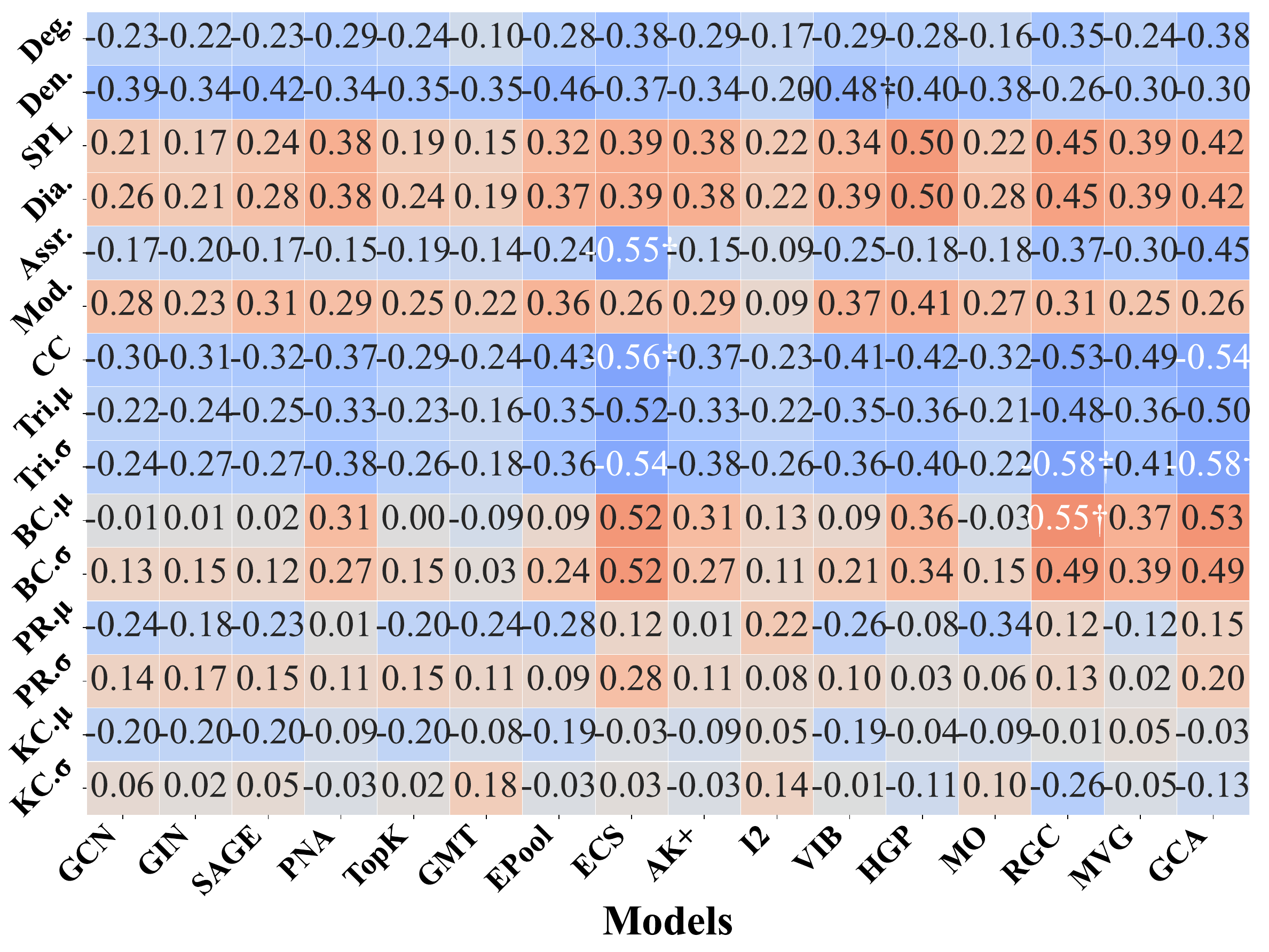}
    
    \vspace{-2pt}

    \includegraphics[width=0.85\linewidth]{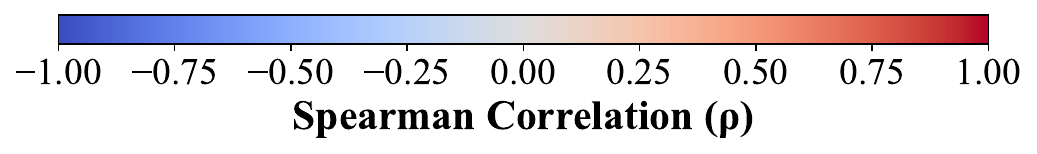}

	\caption{
		Spearman’s $\rho$ between graph features and GNNs performance, with $*$ for $p < 0.05$ and $\dagger$ for $0.05 \le p < 0.1$.
	}
	\label{fig:fig_correlation}
    \vspace{-1em}
\end{figure}

\vspace{1ex}
\noindent\textit{Insights.}
As illustrated in Figure~\ref{fig:fig_correlation}, 
structurally simpler models (e.g., Node-based) show weak correlations with most features, reflecting their reliance on local aggregation rather than global topology. 
Graph density negatively correlates with most models, suggesting high connectivity exacerbates over-smoothing and noise. 
High local clustering and assortativity negatively impact Subgraph-based and SSL-based models, likely due to the ``rich-club'' effect distracting from peripheral structures. 
Conversely, graph sparsity and high Betweenness Centrality positively correlate with Hierarchical (e.g., HGP) and SSL-based models, demonstrating their ability to leverage clear structures for long-range information flow. 
Most importantly, the absence of universal correlations confirms that model selection cannot rely on a single structural indicator, and no architecture consistently dominates. 
This underscores the need for comprehensive benchmarks across diverse domains like \textsc{OpenGLT}, to provide an empirical basis for scenario-specific model selection.

\section{Conclusion and Future Direction}\label{sec:conclusion}

This paper presents a comprehensive experimental study on graph neural networks for graph-level tasks, categorizing existing models, introducing a unified evaluation framework (OpenGLT), and benchmarking 20 GNNs across {\color{black} 26 datasets} under challenging scenarios such as noise, imbalance, and limited data. Our results reveal that no single model excels universally, highlighting important trade-offs between expressiveness and efficiency, and emphasizing the need for robust evaluation in realistic conditions. 

Based on these findings, promising future directions include the development of scenario-adaptive or hybrid GNN architectures that dynamically leverage different model strengths, research into lightweight and scalable algorithms for practical deployment, and the incorporation of transfer and foundation model techniques to improve generalization and data efficiency, particularly when labeled data is scarce. Our study provides valuable benchmarks and guidance for advancing GNN research on graph-level tasks.

\balance

\bibliographystyle{ACM-Reference-Format}
\bibliography{GNN4GC-1}

\end{document}